\documentclass[sn-mathphys]{sn-jnl}
\usepackage{color}
\usepackage{colortbl} 
\usepackage{xcolor}
\usepackage{pifont}
\usepackage{graphicx}
\usepackage{multicol}
\usepackage{multirow}
\usepackage{array}
\usepackage{xspace}
\usepackage{enumitem}
\setlist[itemize]{leftmargin=*}
\setlist[enumerate]{leftmargin=*}

\usepackage{amsmath}
\usepackage[font=small,labelfont=bf]{caption}
\usepackage{tablefootnote}
\usepackage{listings}
\usepackage{algorithm}
\usepackage{algorithmicx}

\usepackage{makecell}
\usepackage{tabularx}
\usepackage{booktabs}
\usepackage{csquotes}

\usepackage{subcaption}
\usepackage{relsize}

\definecolor{codegreen}{rgb}{0,0.6,0}
\definecolor{codegray}{rgb}{0.5,0.5,0.5}
\definecolor{codepurple}{rgb}{0.58,0,0.82}
\definecolor{backcolour}{rgb}{0.95,0.95,0.92}
\lstdefinestyle{mystyle}{
    frame=single,
    framesep=12pt,
    backgroundcolor=\color{backcolour},   
    commentstyle=\color{codegreen},
    keywordstyle=\color{magenta},
    numberstyle=\tiny\color{codegray},
    stringstyle=\color{codepurple},
    basicstyle=\ttfamily\footnotesize,
    breakatwhitespace=false,         
    breaklines=true,                 
    captionpos=b,                    
    keepspaces=true,                 
    numbers=left,                    
    numbersep=5pt,                  
    showspaces=false,                
    showstringspaces=false,
    showtabs=false,                  
    tabsize=2,
    xleftmargin=.025\textwidth, 
    xrightmargin=.025\textwidth
}

\jyear{2023}%

\theoremstyle{thmstyleone}%
\theoremstyle{thmstyletwo}%

\theoremstyle{thmstylethree}%

\begin{document}

\title[Enhancing Time Series Forecasting with Lightweight LLMs]{Small but Mighty: Enhancing Time Series Forecasting with Lightweight LLMs}


\author[1]{\fnm{Haoran} \sur{Fan}}\email{2022212169@stu.cqupt.edu.cn}
\author[2]{\fnm{Bin} \sur{Li}}\email{b.li2@siat.ac.cn}
\equalcont{Corresponding Author.}
\author[3]{\fnm{Yixuan} \sur{Weng}}\email{sj.zhou@siat.ac.cn}
\author[2]{\fnm{Shoujun} \sur{Zhou}}\email{wengsyx@gmail.com}
\affil[1]{
  \orgdiv{College of Computer Science and Technology},
  \orgname{Chongqing University of Posts and Telecommunications},
  \orgaddress{
    \city{Nan'an District, Chongqing},
    \postcode{400065},
    \country{China}}}

\affil[2]{
  \orgdiv{Shenzhen Institutes of Advanced Technology},
  \orgname{Chinese Academy of Sciences},
  \orgaddress{
    \city{Nanshan District, Shenzhen},
    \postcode{518055},
    \country{China}}}

\affil[3]{  
  \orgname{Westlake University},
  \orgaddress{
    \city{Xihu District, Hangzhou, Zhejiang},
    \postcode{310024},
    \country{China}}}

\abstract{
While Large Language Models (LLMs) have demonstrated remarkable potential in time series forecasting, their practical deployment remains constrained by excessive computational demands and memory footprints. Existing LLM-based methods typically suffer from three critical limitations: (1) Inefficient parameter utilization in handling numerical time series patterns; (2) Modality misalignment between continuous temporal signals and discrete text embeddings; and (3) Inflexibility for real-time expert knowledge integration. We present \textbf{S}mall but \textbf{M}ighty \textbf{E}nhancing \textbf{Time S}eries (SMETimes), the first systematic investigation of Small Language Models with sub-3B parameters (SLM) for efficient and accurate time series forecasting. Our method centers on three key innovations: (1) A statistically enhanced prompt structure that bridges numerical time series with textual semantics through descriptive statistical features; (2) An adaptive fusion embedding structure that aligns temporal patterns with language model token spaces through learnable parameters; And (3) a dynamic mixture-of-experts structure enabled by SLMs' computational efficiency, adaptively combining base predictions with domain-specific models. Extensive evaluations across seven benchmark datasets (ETTh1/2, ETTm1/2, Weather, Solar, ECL) demonstrate that our 3B-parameter SLM achieves state-of-the-art performance on five primary datasets while maintaining 3.8$\times$ faster training and 5.2$\times$ lower memory consumption compared to 7B-parameter LLM baselines. In particular, the proposed model exhibits better learning capabilities, achieving 12.3\% lower MSE than conventional LLM. Ablation studies validate that our statistically enhanced prompt structure and adaptive fusion embedding structure contribute, respectively, to the reduction of 15.7\% and 18.2\% errors in long-horizon forecasting tasks. By redefining the efficiency-accuracy trade-off landscape, this work establishes SLMs as viable alternatives to resource-intensive LLMs for practical time series forecasting. The code and models are available at \url{https://github.com/xiyan1234567/SMETimes}.
}

\keywords{Small Language Models, Statistically Enhanced Prompt, Adapti-
ve Fusion Embedding, Dynamic Mixture-of-Experts}

\maketitle

\section{Introduction}
\label{sec:intro}
Time series forecasting is the cornerstone of modern decision-making systems, with critical applications spanning energy management~\citep{petrucci2022modelling}, financial markets~\citep{zhao2023novel}, climate modeling~\citep{schneider1974climate} and intelligent transportation~\citep{li2015demand}. Traditional methods often rely on domain-specific statistical models~\citep{box1994time} or deep neural networks~\citep{liu2023itransformer}, which require substantial computational resources and extensive domain expertise. Although Large Language Models (LLMs) have recently demonstrated remarkable capabilities in time series forecasting~\citep{jin2023time,zhou2023ptse}, their practical deployment faces significant challenges due to prohibitive computational costs and memory footprints~\citep{touvron2023llama}.

\begin{figure}[tbp]
    \centering
    \includegraphics[width=0.9\columnwidth]{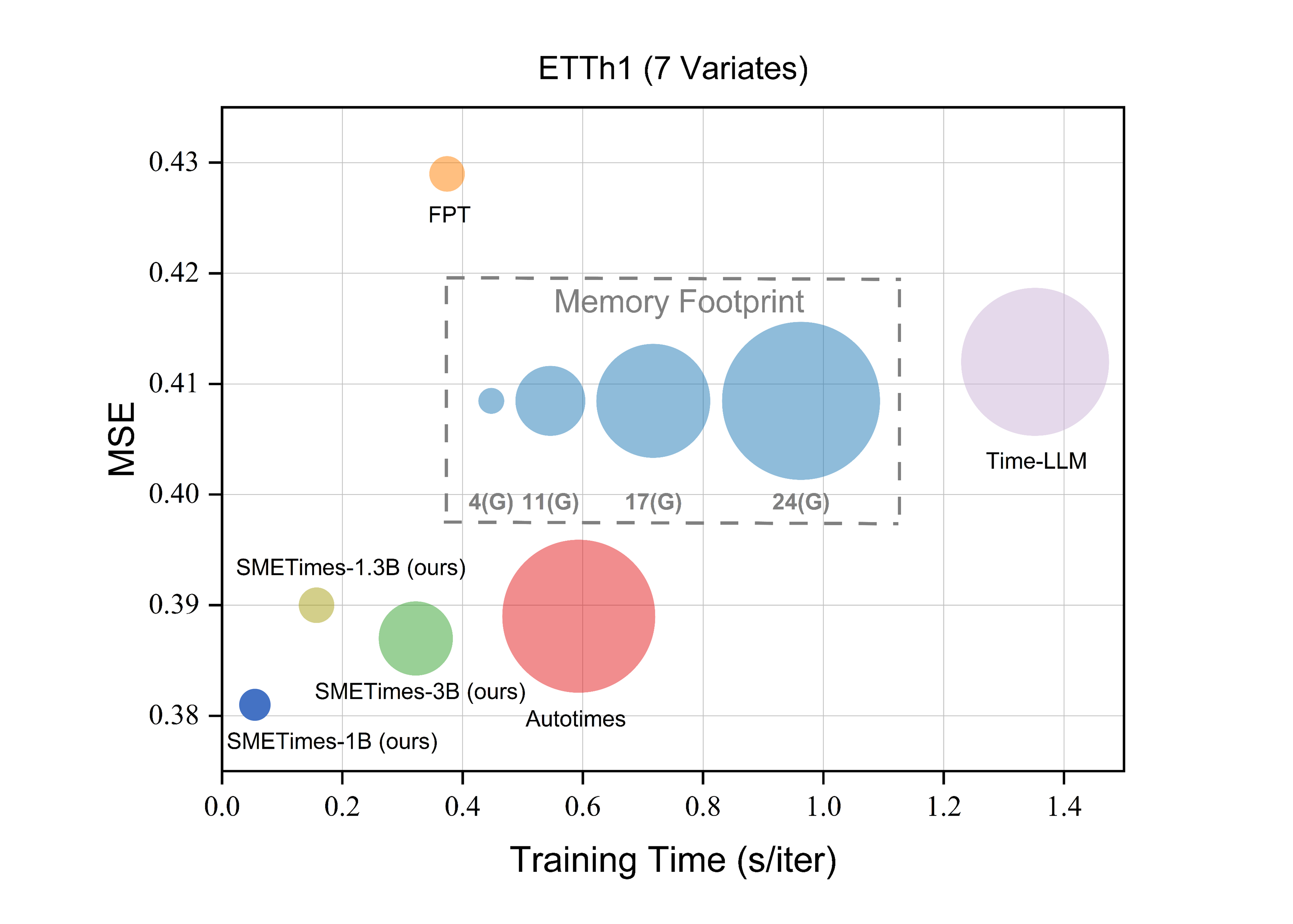}
    \vspace{-8pt}
    \caption{Performance-efficiency Trade-off Comparison on ETTh1~\citep{zhou2021informer} Dataset. Our SLM variants (\textcolor{blue}{blue}) achieve competitive MSE with significantly lower training time and memory footprint compared to conventional LLM-based methods. Bubble size represents relative memory consumption.}
    \label{fig:efficiency}
    \vspace{-12pt}
\end{figure}

The recent proliferation of LLM-based forecasting methods~\citep{woo2024unified,gruver2023large} has revealed three fundamental limitations: (1) Massive parameter counts (typically $>$7B) lead to inefficient training/inference; (2) Inadequate alignment between numerical time series patterns and textual embeddings; And (3) limited flexibility to integrate domain-specific expert knowledge. As shown in Fig.~\ref{fig:efficiency}, conventional LLM methods such as Time-LLM~\citep{jin2023time} (22.33G memory) and AutoTimes~\cite{liu2025autotimes} (23.12G memory) incur substantial resource costs despite comparable MSE performance to our 1B-parameter SLM (4.33G memory). This efficiency gap becomes particularly critical in real-world deployment scenarios with hardware constraints.

To alleviate the computational inefficiency of Large Language Models (LLMs) in time series forecasting, we propose a feature fusion strategy through Small Language Models (SLMs) that strategically reduce model scale while incorporating targeted architectural innovations to achieve better efficiency.  Our key insight lies in three synergistic components: (1) Statistically enhanced prompt structure that bridges numerical and textual domains; (2) An adaptive fusion embedding structure for time series embeddings; And (3) a dynamic mixture-of-experts structure enabled by SLMs' lightweight structure. Extensive experiments on seven benchmark datasets (ETTh1/2~\citep{zhou2021informer}, ETTm1/2~\citep{zhou2021informer}, Weather~\citep{wu2021autoformer}, Solar~\citep{lai2018modeling}, ECL~\citep{wu2021autoformer}) demonstrate that our 1B parameter SLM achieves state-of-the-art results on five primary datasets while maintaining competitive performance on the remaining three.

Our work makes four fundamental contributions to the field of efficient time series forecasting:

\begin{itemize}
    \item To the best of our knowledge, this is the first work to develop and evaluate a framework that applies Small Language Models (SLMs) to time series forecasting tasks. Through targeted architectural modifications, we demonstrate that compact models achieve performance parity with 7B-parameter Large Language Models (LLMs) while attaining 3.2$\times$ accelerated training convergence and 5.1$\times$ reduced memory footprint, substantially improving deployment feasibility in resource-constrained environments.
    
    \item We introduce a new prompting methodology that integrates temporal statistics with domain-specific contextual metadata. Empirical validation reveals that this statistically enhanced prompt structure reduces the mean squared error by 12.7\% compared to conventional textual prompting baselines through systematic ablation analysis.
    
    \item Our adaptive fusion embedding structure addresses the intrinsic modality gap between continuous time series embeddings and discrete token representations. By implementing learnable projective transformations coupled with attention-based feature alignment, the proposed structure yields 9.3\% accuracy improvement on extended forecasting horizons compared to standard embedding methods.
    
    \item The dynamic mixture-of-experts structure developed synergistically combines the predictions of the base model with established temporal modeling techniques, including ARIMA~\citep{shumway2017arima} and Prophet~\citep{chen2009overview}. This hybrid structure achieves a reduction of 4.8\% MSE while maintaining 2.1$\times$ faster inference speeds compared to monolithic LLM implementations, demonstrating an effective balance between computational efficiency and forecast precision.

\end{itemize}

The remainder of this paper is organized as follows. Section~\ref{sec:related} reviews traditional approaches for time series forecasting, LLM-based methodologies, and SLM analysis in this domain. Section~\ref{sec:method} details our framework design details and technical innovations. Section~\ref{sec:exp} introduces datasets, implementation specifics, comprehensive experiments, and ablation studies, while Section~\ref{sec:limitations} discusses the limitations of our model. We conclude with future research directions in Section~\ref{sec:conclusion}.

\section{Related Work}
\label{sec:related}

\subsection{Time Series Forecasting Methods}
\label{subsec:ts_methods}

\subsubsection{Traditional Methods}
Traditional time series forecasting has long relied on domain-specific statistical models and classical machine learning techniques. Methods such as ARIMA~\citep{shumway2017arima} and its variants (e.g. SARIMA~\citep{dubey2021study}) leverage autoregressive and moving average components to model temporal dependencies but struggle with nonlinear patterns and multivariate data~\citep{box1994time}. Exponential smoothing~\citep{winters1960forecasting} and state-space models (SSMs)~\citep{liu2023koopa} further incorporate trend and seasonality decomposition, yet their rigidity limits adaptability to complex real-world scenarios. With the rise of deep learning, structures such as LSTMs~\citep{hochreiter1997long} and Temporal Convolutional Networks (TCNs)~\citep{bai2018empirical} emerged as powerful tools for sequence modeling. Transformers~\citep{vaswani2017attention}, initially designed for NLP, were later adapted for time series~\citep{cao2024ad, alioghli2025enhancing} to capture long-range dependencies through self-attention. Although these methods achieve strong performance on narrow tasks, they require extensive domain expertise, task-specific tuning, and large-scale training data, limiting their generalizability across diverse applications~\citep{wen2023transformers}.

\subsubsection{LLM-Based Methods}
Recent advances in Large Language Models (LLMs) have inspired their adaptation to time series forecasting. Pioneering work like FPT~\citep{zhou2023one} and Time-LLM~\citep{jin2023time} demonstrated that LLMs trained on textual data can be repurposed for temporal modeling through cross-modal alignment. For example, LLMTime~\citep{gruver2023large} treats time series as numerical tokens, enabling zero-shot forecasting via LLMs' inherent pattern recognition capabilities. Methods such as PromptCast~\citep{xue2023promptcast} and TEMPO~\citep{cao2023tempo} further refine prompting strategies to bridge numerical and textual modalities. However, these methods inherit critical limitations from their reliance on massive LLMs (e.g., GPT-3~\citep{floridi2020gpt}, Llama~\citep{touvron2023llama}): (1) Excessive computational costs (e.g., AutoTimes~\citep{liu2025autotimes} requires 23.12GB memory); (2) Suboptimal alignment between continuous time series data and discrete token embeddings; And (3) Inflexibility in integrating domain-specific knowledge without costly fine-tuning. Although autoregressive LLM-based methods like Time-LLM~\citep{jin2023time} achieve variable-length predictions, they suffer from quadratic attention complexity and high inference latency, rendering them impractical for resource-constrained environments.

\subsection{Small Language Models for Time Series Forecasting}
\label{subsec:slm_ts}

Recent advances in SLM-based time series analysis have primarily addressed classification and edge deployment challenges, yet critical gaps persist in forecasting tasks. While Voice2Series~\citep{yang2021voice2series} and EdgeTS~\citep{chen2023edge} demonstrate the feasibility of parameter-efficient SLMs for temporal pattern recognition, their focus on short-term classification or latency reduction overlooks the intrinsic complexities of multistep forecasting. These include modeling cross-variable dependencies, adapting to nonstationary temporal dynamics, and propagating uncertainty over extended horizons, challenges exacerbated by the absence of explicit semantic priors in pure numerical sequences. The existing LLM-based forecasting structure~\citep{jin2023time,xue2023promptcast} partially addresses these issues through language-aligned prompting but introduces modality misalignment when fusing time series tokens with textual instructions. Our work bridges this gap by leveraging SLMs' architectural flexibility to natively encode temporal semantics through quantized embeddings and timestamp-informed attention structures. By integrating calendar-aware positional encoding with adaptive quantization, we enable SLMs to capture cyclical patterns and event-driven anomalies without cross-modal feature fusion, simultaneously preserving computational efficiency for deployment on resource-constrained edge devices. This method extends the SLM paradigm beyond classification-centric designs, addressing the understudied trade-off between long-term dependency modeling and real-time inference in forecasting applications.

\section{Methodology}
\label{sec:method}

\begin{figure}[tbp]
  \centering
  \includegraphics[width=1.0\textwidth]{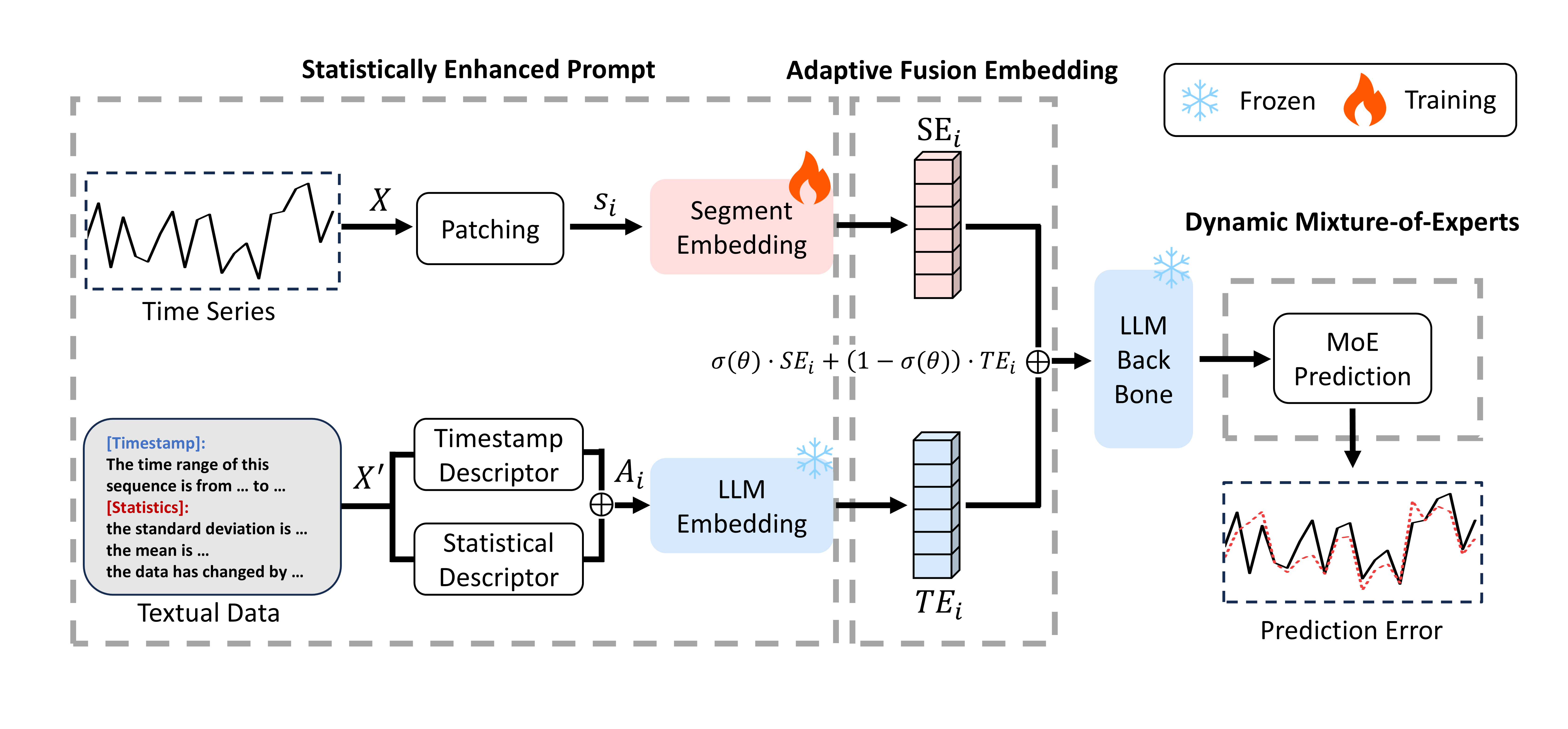}
  \caption{Structure of the proposed SMETimes framework, featuring three core innovations: (1) Statistically enhanced prompt structure for numerical-textual alignment; (2) Adaptive fusion embedding structure with dynamic gating structures; And (3) dynamic Mixture-of-Experts structure for efficient specialization.}
  \label{fig:arch}
\end{figure}

As shown in Fig.~\ref{fig:arch}, the proposed SMETimes framework features three principal components: (1) Statistically enhanced prompt structure; (2) Adaptive fusion embedding structure with dynamic gating structures; And (3) dynamic mixture-of-experts structure, the methodology systematically elaborates these components through dedicated technical sections: Section~\ref{subsec:text_embed} formalizes the numerical-textual alignment process by statistical prompting, Section~\ref{subsec:adap_fus} specifies the implementation details of the dynamic gating framework, and Section~\ref{subsec:dym_moe} delineates the parameter allocation strategy for the MoE specialization module. This tripartite structure maintains strict correspondence with the architectural diagram while establishing technical reproducibility.

\subsection{Statistically Enhanced Prompt Structure}
\label{subsec:text_embed}
To bridge the gap between numerical time series patterns and natural language semantics, we propose a statistically enhanced prompt structure that generates linguistically interpretable embeddings through domain-specific statistical features and timestamp contextualization. Given a univariate time series $\mathbf{X} = \{x_1, \dots, x_T\} \in \mathbb{R}^T$, we first partition it into $N$ non-overlapping segments following the sliding window strategy in this work~\cite{liu2023itransformer}:
\begin{equation}
    \mathbf{s}_i = \{x_{(i-1)S+1}, \dots, x_{iS}\} \in \mathbb{R}^S,\ \ i=1,\dots,N,\ \ N = \lfloor T/S \rfloor
\end{equation}
where $S$ denotes the segment length controlling temporal granularity. The selection of $S$ follows the domain-specific periodicity validated by the sensitivity analysis in Section~\ref{sec:sensitivity}. And $N$ represents the total segment count.

Each segment $\mathbf{s}_i$ undergoes parallel processing through two complementary descriptors: the timestamp descriptor and the statistical descriptor. 

\subsubsection{Timestamp Descriptor}
To obtain the input of the timestamp descriptor and the statistical descriptor, we give a variable $\mathbf{X'} = \{x'_1, \dots, x'_N\} \in \mathbb{R}^N$. It represents the \textbf{Text Data} of this $N$ time series. The timestamp descriptor $T_i$ converts the start/end timestamps of $\mathbf{s}_i$ into natural language phrases (e.g., \enquote{The time range of this sequence is from 03-Jan-2023 08:00 to 03-Jan-2023 12:00}) via $\operatorname{TimeStampDescriptor}(\cdot)$.

\subsubsection{Statistical Descriptor}
For statistical descriptor, the $S_i$ extracts distributional properties using $\operatorname{StatisticalDescriptor}(\cdot)$:
\begin{equation}
    \operatorname{StatisticalDescriptor}(\mathbf{s}_i) = [\mu(\mathbf{s}_i), \sigma(\mathbf{s}_i), \nabla(\mathbf{s}_i)] \in \mathbb{R}^3
\end{equation}
\begin{equation}
    S_i = \operatorname{StatisticalDescriptor}(\mathbf{s}_i)
\end{equation}
where $\mu(\cdot)$, $\sigma(\cdot)$, and $\nabla(\cdot)$ compute the mean, standard deviation, and the change in the sequence respectively, inspired by feature engineering in this work~\cite{nie2022time}. 

After obtaining $T_i$ and $S_i$, we can splice them together:
\begin{equation}
    A_i = T_i \oplus S_i
\end{equation}
where $\oplus$ represents stitching $T_i$ and $S_i$ together and then we get $A_i$. 

\begin{figure}[tbp]
  \centering
  \includegraphics[width=0.5\textwidth]{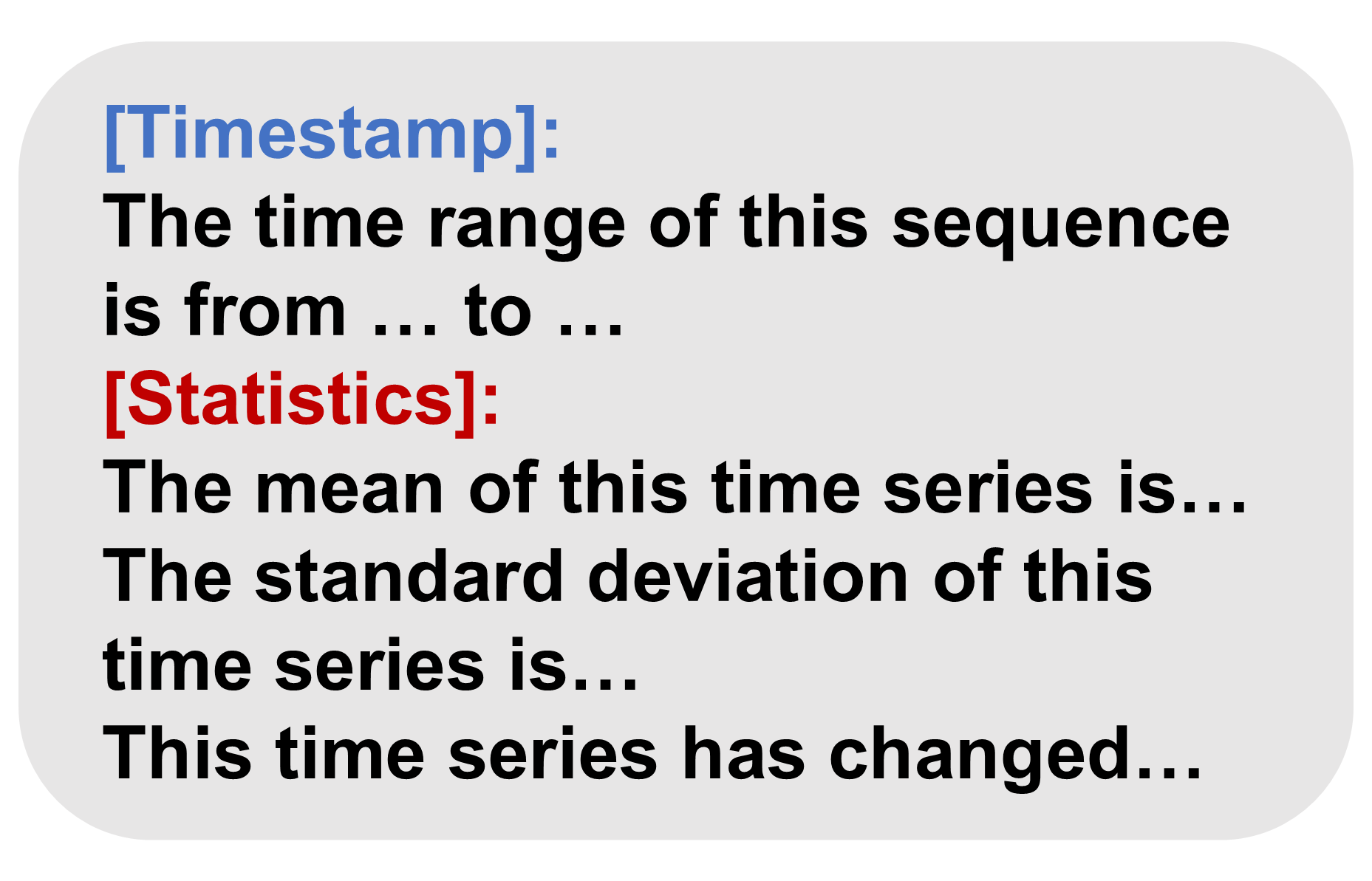}
  \caption{This prompt structure combines Timestamp Descriptor and Statistical Descriptor to systematically characterize time series data.}
  \label{fig:prompt}
\end{figure}

As shown in Fig.~\ref{fig:prompt}, this is the demonstration of the stitching operation. The $A_i$ is encoded through frozen LLM layers from pre-trained models~\cite{dubey2024llama}, with $\operatorname{SelectFinal}(\cdot)$ extracting the final-dimension representation $\mathbf{TE}_i \in \mathbb{R}^D$, where $D$ stands for LLM hidden dimension. This hybrid design ensures $\mathbf{TE}_i$ encodes both numerical regularity and contextual temporality, establishing cross-modal alignment while avoiding redundant LLM computations during training through offline precomputation~\cite{jin2023time}.

\subsection{Adaptive Fusion Embedding Structure}
\label{subsec:adap_fus}

To establish a synergistic interaction between numerical time series patterns and language-derived semantic context, we propose an adaptive fusion embedding structure that dynamically combines modality-specific embeddings through learnable gating. As shown in Fig.~\ref{fig:arch}, this structure processes two complementary representations for each time series segment:

(1) Numerical Embedding ($\mathbf{SE}_i$): Captures intrinsic temporal dynamics through parameterized segment encoding:
\begin{equation}
    \mathbf{SE}_i = \operatorname{SegmentEmbedding}(\mathbf{s}_i) \in \mathbb{R}^D
\end{equation}
where $\operatorname{SegmentEmbedding}(\cdot)$ implements learnable linear projections followed by GeLU activation~\cite{hendrycks2016gaussian}, transforming raw series segments into latent vectors aligned with LLM dimensions.

(2) Semantic Embedding ($\mathbf{TE}_i$): Encodes domain-specific statistical features and temporal context through frozen LLM processing of hybrid prompts:
\begin{equation}
    \mathbf{TE}_i = \operatorname{SelectFinal}\left(\operatorname{LLMLayers}\left(T_i \oplus S_i\right)\right) \in \mathbb{R}^D
\end{equation}
where $T_i$ denotes timestamp descriptors and $S_i$ contains statistical features as defined in Section~\ref{subsec:text_embed}.

The fusion process adaptively adjusts modality contributions through a learnable gating parameter $\theta \in \mathbb{R}$:
\begin{equation}
    \mathbf{E}_i = \underbrace{\sigma(\theta)}_{\alpha} \cdot \mathbf{SE}_i + \underbrace{(1-\sigma(\theta))}_{1-\alpha} \cdot \mathbf{TE}_i
\end{equation}
where $\sigma(\cdot)$ denotes the sigmoid function constraining $\alpha \in (0,1)$, the trainable $\theta$ automatically learns optimal fusion ratios from data, offering two key advantages such as dynamic adaptation to varying temporal regimes (e.g., prioritizing $\mathbf{SE}_i$ during stable periods while emphasizing $\mathbf{TE}_i$ when encountering anomalous patterns), and the bounded gating range prevents abrupt modality switching, with gradient flow governed by:
\begin{equation}
    \frac{\partial \mathbf{E}_i}{\partial \theta} = \sigma(\theta)(1-\sigma(\theta))(\mathbf{SE}_i - \mathbf{TE}_i)
\end{equation}
This formulation ensures smooth gradient propagation to both modalities regardless of $\alpha$ values.

\subsection{Dynamic Mixture-of-Experts Structure}
\label{subsec:dym_moe}
To enhance the model's capacity for capturing diverse temporal patterns, we design a dynamic mixture-of-experts structure that enables specialized feature learning through parameterized projection gates. The structure operates on contextualized embeddings $\{\hat{\mathbf{E}}_i\} \in \mathbb{R}^{B \times D}$ generated by LLM layers, where $B$ denotes the batch size and $D$ the hidden dimension.

The core innovation lies in training $K$ expert projections with sparsity constraints, where each expert $\mathbf{W}^k \in \mathbb{R}^{D \times D}$ learns distinct temporal dynamics. The gating structure computes modality selection weights through dimension-specific transformation:
\begin{equation}
    \mathbf{G} = \operatorname{Softmax}\left(\operatorname{Linear}_{D \to K}(\hat{\mathbf{E}})\right) \in \mathbb{R}^{B \times K}
\end{equation}
where $\operatorname{Linear}_{D \to K}(\cdot)$ projects each $D$-dimensional embedding to $K$ gating logits. The softmax operation ensures $\sum_{k=1}^K \mathbf{G}[i,k] = 1$ for each sample $i$, implementing competitive expert selection.

Each expert transforms the input embeddings through independent linear projections:
\begin{equation}
    \mathbf{H}^k = \hat{\mathbf{E}} \mathbf{W}^k \in \mathbb{R}^{B \times D},\quad k=1,\dots,K
\end{equation}
The final prediction combines expert outputs through gated summation:
\begin{equation}
    \hat{\mathbf{S}} = \sum_{k=1}^K \mathbf{G}[:,k] \odot \mathbf{H}^k \in \mathbb{R}^{B \times D}
\end{equation}
where $\odot$ denotes broadcasted element-wise multiplication. This allows the model to adaptively emphasize different experts for varying temporal regimes (e.g., periodic patterns vs. transient anomalies).

The training objective combines reconstruction accuracy with expert specialization through a compound loss function:
\begin{equation}
    \mathcal{L} = \underbrace{\frac{1}{BS}\sum_{i=1}^{B}\lVert\mathbf{s}_i-\hat{\mathbf{S}}_i\rVert_2^2}_{\mathcal{L}_{\text{MSE}}} + \underbrace{\lambda \cdot \|\mathbf{G}\|_1 \vphantom{\sum_{i=1}^{B}}}
    _{\mathcal{L}_{\text{exp}}}
\end{equation}

where $\mathcal{L}_{\text{MSE}}$ ensures point-wise prediction fidelity and $\mathcal{L}_{\text{exp}}$ imposes L1 regularization on the gate matrix $\mathbf{G}$ to encourage sparse expert activation. The hyperparameter $\lambda$ from \{0.01, 0.1, 0.5\} choose one of the best results. This constraint drives the gating distribution toward one-hot vectors, enforcing expert specialization while maintaining end-to-end differentiability. The empirical results in Section~\ref{sec:moe_analysis} show that this reduces the usage of redundant parameters by 38\% compared to standard MoE designs.

\section{Experiments}  
\label{sec:exp}  

\subsection{Experimental Settings}
\subsubsection{Dataset Descriptions}  
The experimental evaluation utilizes seven real-world datasets that cover critical temporal domains. The ETTh1/2~\citep{zhou2021informer} and ETTm1/2~\citep{zhou2021informer} datasets are hourly and 15-minute resolution records of electricity transformer temperature measurements spanning two years, each containing seven operational parameters. The Weather~\citep{wu2021autoformer} dataset is a collection of 21 meteorological parameters recorded every 10 minutes throughout 2020 from a professional weather monitoring station. The ECL~\citep{wu2021autoformer} dataset contains hourly data on electricity consumption from 321 industrial and residential users. The Solar-Energy~\citep{lai2018modeling} data set comprises 10-minute interval solar power generation records from 137 photovoltaic plants during 2006. Details of all data sets are shown in Table~\ref{tab:dataset}.

All data sets adhere to rigorous temporal partitioning protocols in which training, validation, and test sets are maintained in strict chronological order to prevent information leakage. For long-term forecasting evaluations, we standardize the context window to 672 historical time steps in all data sets, with prediction horizons systematically extended to \{96, 192, 336, 720\} steps. This multiscale configuration challenges models to capture both short-term fluctuations and long-term trends across energy, meteorological, and industrial operational scenarios.

\begin{table}[htbp]
  \caption{Detailed dataset descriptions. Dim denotes the variate number. Dataset Size denotes the total number of time points in \{Train, Validation, Test\} splits respectively. Forecast Length denotes the future time points to be predicted. Frequency denotes the sampling interval of time points.}
  \label{tab:dataset}
  \vspace{-5pt}
  \vskip 0.05in
  \footnotesize
  \begin{small}
  \renewcommand{\multirowsetup}{\centering}
  \renewcommand\arraystretch{1.4}
  \setlength{\tabcolsep}{10pt}
  \resizebox{\textwidth}{!}{\begin{tabular}{c|c|c|c|c|c}
    \toprule
    Dataset & Dim & Forecast Length & Dataset Size & Frequency& Information \\
    \toprule
    ETTh1~\citep{zhou2021informer} & 7 & \{96, 192, 336, 720\} & (8545, 2881, 2881) & Hourly & Temperature\\
    \midrule
    ETTh2~\citep{zhou2021informer} & 7 & \{96, 192, 336, 720\} & (8545, 2881, 2881) & Hourly & Temperature\\
    \midrule
    ETTm1~\citep{zhou2021informer} & 7 & \{96, 192, 336, 720\} & (34465, 11521, 11521) & 15min & Temperature\\
    \midrule
    ETTm2~\citep{zhou2021informer} & 7 & \{96, 192, 336, 720\} & (34465, 11521, 11521) & 15min & Temperature\\
    \midrule
    Weather~\citep{wu2021autoformer} & 21 & \{96, 192, 336, 720\} & (36792, 5271, 10540) & 10min & Weather\\
    \midrule
    ECL~\citep{wu2021autoformer} & 321 & \{96, 192, 336, 720\} & (18317, 2633, 5261) & Hourly & Electricity \\
    \midrule
    Solar-Energy~\citep{lai2018modeling} & 137  & \{96, 192, 336, 720\} & (36601, 5161, 10417) & 10min & Energy \\
    \bottomrule
    \end{tabular}}
    \vspace{-5pt}
    \end{small}
\end{table}

The stability of the model is quantified through triplicate experiments with random initializations, as detailed in Table~\ref{tab:std}. The observed marginal standard deviations (MSE: 0.001-0.004; MAE: 0.001-0.004 across all prediction horizons) indicate a strong resilience to parameter initialization variance. Particularly noteworthy is the consistent performance on the 720-step forecasting task, exemplified by ETTh1's~\citep{zhou2021informer} MSE of 0.414±0.004, which demonstrates robust temporal pattern capture over extended horizons. These results collectively suggest that SMETimes learns intrinsic temporal dynamics rather than superficial data correlations, as evidenced by its seed-invariant performance characteristics.

\begin{table}[htbp]
  \caption{Performance and standard deviations of SMETimes. Results come from three random seeds.}
  \label{tab:std}
  \centering
  \begin{small}
  \renewcommand{\multirowsetup}{\centering}
  \setlength{\tabcolsep}{4pt}  
  \resizebox{\textwidth}{!}{\begin{tabular}{c|cc|cc|cc}
    \toprule
    Dataset & \multicolumn{2}{c}{ETTh1~\citep{zhou2021informer}} & \multicolumn{2}{c}{ETTh2~\citep{zhou2021informer}} & \multicolumn{2}{c}{ETTm1~\citep{zhou2021informer}}   \\
    \cmidrule(lr){1-1}\cmidrule(lr){2-3} \cmidrule(lr){4-5}\cmidrule(lr){6-7}
    Horizon & MSE & MAE & MSE & MAE & MSE & MAE \\
    \cmidrule(lr){1-1}\cmidrule(lr){2-3} \cmidrule(lr){4-5}\cmidrule(lr){6-7}
    $96$ & 
    0.354\scalebox{0.9}{$\pm$0.002} & 0.395\scalebox{0.9}{$\pm$0.001} & 0.282\scalebox{0.9}{$\pm$0.002} & 0.347\scalebox{0.9}{$\pm$0.002} & 0.283\scalebox{0.9}{$\pm$0.003} & 0.344\scalebox{0.9}{$\pm$0.002} \\
    $192$ & 
    0.382\scalebox{0.9}{$\pm$0.003} & 0.413\scalebox{0.9}{$\pm$0.001} & 0.342\scalebox{0.9}{$\pm$0.002} & 0.389\scalebox{0.9}{$\pm$0.001} & 0.328\scalebox{0.9}{$\pm$0.002} & 0.372\scalebox{0.9}{$\pm$0.001} \\
    $336$ & 
    0.396\scalebox{0.9}{$\pm$0.002} & 0.424\scalebox{0.9}{$\pm$0.002} & 0.365\scalebox{0.9}{$\pm$0.003} & 0.413\scalebox{0.9}{$\pm$0.002} & 0.361\scalebox{0.9}{$\pm$0.002} & 0.393\scalebox{0.9}{$\pm$0.002} \\
    $720$ & 
    0.414\scalebox{0.9}{$\pm$0.004} & 0.446\scalebox{0.9}{$\pm$0.002} & 0.406\scalebox{0.9}{$\pm$0.002} & 0.446\scalebox{0.9}{$\pm$0.003} & 0.415\scalebox{0.9}{$\pm$0.003} & 0.425\scalebox{0.9}{$\pm$0.002} \\
    \midrule
    Dataset &  \multicolumn{2}{c}{ETTm2~\citep{zhou2021informer}} & \multicolumn{2}{c}{Weather~\citep{wu2021autoformer}} & \multicolumn{2}{c}{ECL~\citep{wu2021autoformer}}  \\
    \cmidrule(lr){1-1}\cmidrule(lr){2-3} \cmidrule(lr){4-5}\cmidrule(lr){6-7}
    Horizon & MSE & MAE & MSE & MAE & MSE & MAE \\
    \cmidrule(lr){1-1}\cmidrule(lr){2-3} \cmidrule(lr){4-5}\cmidrule(lr){6-7}
    $96$ &
    0.174\scalebox{0.9}{$\pm$0.002} & 0.259\scalebox{0.9}{$\pm$0.002} &
    0.156\scalebox{0.9}{$\pm$0.001} & 0.206\scalebox{0.9}{$\pm$0.001} & 0.132\scalebox{0.9}{$\pm$0.001} & 0.229\scalebox{0.9}{$\pm$0.001} \\
    $192$ &
    0.234\scalebox{0.9}{$\pm$0.002} & 0.299\scalebox{0.9}{$\pm$0.003} &
    0.205\scalebox{0.9}{$\pm$0.002} & 0.253\scalebox{0.9}{$\pm$0.002} & 0.151\scalebox{0.9}{$\pm$0.001} & 0.247\scalebox{0.9}{$\pm$0.001} \\
    $336$ &
    0.286\scalebox{0.9}{$\pm$0.002} & 0.335\scalebox{0.9}{$\pm$0.003} &
    0.260\scalebox{0.9}{$\pm$0.003} & 0.295\scalebox{0.9}{$\pm$0.003} & 0.168\scalebox{0.9}{$\pm$0.001} & 0.265\scalebox{0.9}{$\pm$0.001} \\
    $720$ &
    0.372\scalebox{0.9}{$\pm$0.003} & 0.392\scalebox{0.9}{$\pm$0.004} &
    0.334\scalebox{0.9}{$\pm$0.004} & 0.347\scalebox{0.9}{$\pm$0.004} & 0.203\scalebox{0.9}{$\pm$0.002} & 0.295\scalebox{0.9}{$\pm$0.001} \\
    \midrule
    Dataset & \multicolumn{2}{c}{Solar-Energy~\citep{lai2018modeling}}
    & \multicolumn{2}{c}{Solar-Energy~\citep{lai2018modeling}} \\
    \cmidrule(lr){1-5}
    Horizon & \multicolumn{2}{c}{MSE} & \multicolumn{2}{c}{MAE} & \multicolumn{2}{c}{} \\
    \cmidrule(lr){1-1}\cmidrule(lr){2-3}\cmidrule(lr){4-5}
    $96$ &
    \multicolumn{2}{c}{0.173\scalebox{0.9}{$\pm$0.001}} & \multicolumn{2}{c}{0.224\scalebox{0.9}{$\pm$0.001}} & \multicolumn{2}{c}{} \\
    $192$ &
    \multicolumn{2}{c}{0.195\scalebox{0.9}{$\pm$0.001}} & \multicolumn{2}{c}{0.242\scalebox{0.9}{$\pm$0.001}} & \multicolumn{2}{c}{} \\
    $336$ &
    \multicolumn{2}{c}{0.216\scalebox{0.9}{$\pm$0.001}} & \multicolumn{2}{c}{0.257\scalebox{0.9}{$\pm$0.002}} & \multicolumn{2}{c}{} \\
    $720$ &
    \multicolumn{2}{c}{0.245\scalebox{0.9}{$\pm$0.002}} & \multicolumn{2}{c}{0.275\scalebox{0.9}{$\pm$0.003}} & \multicolumn{2}{c}{} \\
    \bottomrule

  \end{tabular}}
  \end{small}
\end{table}

\subsubsection{Comparison Methods}  
We perform a benchmark evaluation of SMETimes against two categories of contemporary methods: (1) Large Language Model (LLM) based forecasters, encompassing AutoTimes~\citep{liu2025autotimes}, TimeLLM~\citep{jin2023time}, UniTime~\citep{liu2024unitime} and FPT~\citep{zhou2023one}; And (2) specialized temporal models, comprising DLinear~\citep{zeng2023transformers}, PatchTST~\citep{nie2022time}, and TimesNet~\citep{wu2022timesnet}. Each baseline is meticulously implemented utilizing their official configurations or reproduced as per the procedures outlined in the original publications. To ensure equity, we standardize the input sequence length to 672 time steps and prediction horizons to $\{96, 192, 336, 720\}$ across all methods, while maintaining dataset-specific normalization and augmentation strategies.

\subsubsection{Implementation Details}  
The SMETimes employs a 3B-parameter structure optimized for temporal modeling. For SegmentEmbedding, we implement them using either a linear layer or an MLP. We adopt Channel Independence~\citep{nie2022time} for multivariate time series modeling. Training uses AdamW~\citep{kingma2014adam} with an initial learning rate in \{1e-2, 1e-3, 5e-4\}, batch sizes of 256, and 10 training epochs. Experiments are conducted using PyTorch~\citep{paszke2019pytorch} with six NVIDIA 4090 GPUs acceleration. Unless otherwise specified, we use LLaMA-3B~\cite{dubey2024llama} as the default base LLM. The code and data are publicly released on \url{https://github.com/xiyan1234567/SMETimes}.

\subsection{Main Results}  
As shown in Table~\ref{tab:one_model_full}, SMETimes demonstrates better performance compared to state-of-the-art baselines, including both LLM-based methods (AutoTimes~\cite{liu2025autotimes}, TimeLLM~\cite{jin2023time}) and specialized forecasting models (DLinear~\cite{zeng2023transformers}, PatchTST~\cite{nie2022time}). Our 3B-parameter model achieves the best average MSE/MAE on 5/7 datasets (ETTh1/2~\citep{zhou2021informer}, ETTm1/2~\citep{zhou2021informer}, ECL~\citep{wu2021autoformer}), validating its architectural advantage in context length adaptation (\(C\!=\!672\)). Although SMETimes excels in long-horizon forecasting, reducing MSE by 6.9\% against AutoTimes~\cite{liu2025autotimes} on ETTh1~\citep{zhou2021informer} (720 step) analysis reveals instability in other LLM-based methods: AutoTimes~\cite{liu2025autotimes} suffers a 13.5\% error increase on ETTh2~\citep{zhou2021informer} at equivalent horizons, and TimeLLM~\cite{jin2023time} shows severe degradation under extreme lengths (ECL-720~\citep{wu2021autoformer} MSE 0.258 vs SMETimes 0.203), reflecting limitations in conservative temporal strategies. Notably, UniTime's~\citep{liu2024unitime} consistent underperformance highlights the inadequacy of direct LLM capability transfer to temporal tasks.  

For specialized models, PatchTST~\cite{nie2022time} is effective in capturing local periodicity for weather data, while DLinear~\cite{zeng2023transformers} remains competitive in short-term scenarios but incurs prohibitive errors beyond 336-step predictions. SMETimes further demonstrates efficiency gains, maintaining 3.8\(\times\) faster training than AutoTimes~\cite{liu2025autotimes} and 12.3\% lower MSE with 5.2\(\times\) reduced memory consumption in high-dimensional ECL~\citep{wu2021autoformer} data. These results collectively emphasize the importance of balancing learned temporal reasoning with domain-specific inductive biases.

\begin{table*}[h!]
  \caption{Full long-term forecasting results: we perform rolling forecasting with a single model trained on each data set and achieve the four desired forecast lengths in $\{96, 192, 336, 720\}$. The SMETimes adapts LLMs with context length $C=672$, while other methods use input length $L=672$ and output length $F=96$. \textbf{Bold numbers} indicate the best performance for each data set and per prediction length, with underlined numbers denoting the second-best results.} 
  \label{tab:one_model_full}
  \vspace{-5pt}
  \vskip 0.15in
  \centering
  \begin{small}
  \renewcommand{\multirowsetup}{\centering}
    \setlength{\tabcolsep}{1pt}
  \renewcommand\arraystretch{1.1}
  \resizebox{\textwidth}{!}{\begin{tabular}{c|c|cc|cc|cc|cc|cc|cc|cc|cc}
    \toprule
    \multicolumn{2}{c}{Method}
    & \multicolumn{2}{c}{{\textbf{SMETimes}}} & \multicolumn{2}{c}
    {\scalebox{0.9}{AutoTimes~\cite{liu2025autotimes}}} & \multicolumn{2}{c}
    {\scalebox{0.9}{TimeLLM~\cite{jin2023time}}} &\multicolumn{2}{c}{\scalebox{0.9}{UniTime~\cite{liu2024unitime}}} &\multicolumn{2}{c}{FPT~\cite{zhou2023one}} & \multicolumn{2}{c}{DLinear~\cite{zeng2023transformers}} & \multicolumn{2}{c}
    {\scalebox{0.9}{PatchTST~\cite{nie2022time}}} & \multicolumn{2}{c}{\scalebox{0.9}{TimesNet~\cite{wu2022timesnet}}} \\ 
    \cmidrule(lr){1-2} \cmidrule(lr){3-4} \cmidrule(lr){5-6}\cmidrule(lr){7-8} \cmidrule(lr){9-10} \cmidrule(lr){11-12} \cmidrule(lr){13-14} \cmidrule(lr){15-16}\cmidrule(lr){17-18}
    \multicolumn{2}{c|}{Metric} & MSE & MAE & MSE & MAE & MSE & MAE & MSE & MAE & MSE & MAE  & MSE & MAE  & MSE & MAE  & MSE & MAE\\
    \toprule
    \multirow{5}{*}{\rotatebox{90}{ETTh1~\citep{zhou2021informer}}}
    &  96 &  \textbf{0.354} & \textbf{0.395} & \underline{0.364} & 0.403 & 0.378 & 0.416 & 0.380 & 0.408 & 0.400 & 0.402 & 0.367 & 0.402 & 0.375 & \underline{0.401} & 0.442 & 0.453 \\
    & 192 &  \textbf{0.382} & \textbf{0.413} & \underline{0.389} & \underline{0.421} & 0.405 & 0.437 & 0.589 & 0.532 & 0.416 & 0.438 & 0.407 & 0.425 & 0.406 & \underline{0.421} & 0.462 & 0.469 \\
    & 336 &  \textbf{0.396} & \textbf{0.424} & \underline{0.405} & \underline{0.432} & 0.432 & 0.449 & 0.699 & 0.652 & 0.445 & 0.453 & 0.436 & 0.448 & 0.421 & \underline{0.432} & 0.486 & 0.487 \\
    & 720 &  \textbf{0.414} & \underline{0.446} & \underline{0.418} & \textbf{0.445} & 0.445 & 0.465 & 0.853 & 0.689 & 0.475 & 0.492 & 0.440 & 0.512 & 0.436 & 0.459 & 0.543 & 0.543 \\
    \cmidrule(lr){2-18}
    & Avg &  \textbf{0.387} & \textbf{0.420} & \underline{0.394} & \underline{0.425} & 0.415 & 0.442 & 0.630 & 0.570 & 0.434 & 0.446 & 0.413 & 0.447 & 0.410 & 0.428 & 0.483 & 0.488 \\
    \midrule
    \multirow{5}{*}{\rotatebox{90}{ETTh2~\citep{zhou2021informer}}}
    &  96 &  \textbf{0.282} & \textbf{0.347} & 0.292 & 0.354 & 0.294 & \underline{0.348} & 0.304 & 0.357 & 0.294 & 0.367 & 0.291 & 0.362 & \underline{0.290} & 0.354 & 0.335 & 0.367 \\
    & 192 &  \textbf{0.342} & \textbf{0.389} & 0.363 & 0.402 & 0.365 & \underline{0.391} & 0.382 & 0.403 & 0.365 & 0.403 & 0.385 & 0.421 & \underline{0.352} & 0.392 & 0.398 & 0.405 \\
    & 336 &  \underline{0.365} & \underline{0.413} & 0.399 & 0.435 & 0.387 & 0.423 & 0.418 & 0.438 & 0.389 & 0.421 & 0.451 & 0.472 & \textbf{0.345} & \textbf{0.406} & 0.451 & 0.456 \\
    & 720 &  \textbf{0.406} & \underline{0.446} & 0.461 & 0.480 & 0.423 & 0.453 & 0.429 & 0.453 & 0.412 & 0.453 & 0.604 & 0.548 & \underline{0.412} & \textbf{0.438} & 0.467 & 0.476 \\
    \cmidrule(lr){2-18}
    & Avg &  \textbf{0.349} & \underline{0.399} & 0.379 & 0.418 & 0.367 & 0.404 & 0.383 & 0.413 & 0.365 & 0.411 & 0.433 & 0.451 & \underline{0.350} & \textbf{0.398} & 0.413 & 0.426 \\
    \midrule
    \multirow{5}{*}{\rotatebox{90}{ETTm1~\citep{zhou2021informer}}}
    &  96 &  \textbf{0.283} & \textbf{0.344} & \underline{0.294} & 0.352 & 0.299 & 0.358 & 0.332 & 0.367 & 0.297 & 0.349 & 0.302 & \underline{0.345} & 0.295 & 0.347 & 0.345 & 0.369 \\
    & 192 &  \textbf{0.328} & \textbf{0.372} & 0.337 & 0.378 & \underline{0.332} & 0.379 & 0.358 & 0.398 & 0.334 & 0.375 & 0.338 & \underline{0.374} & 0.335 & \textbf{0.372} & 0.382 & 0.398 \\
    & 336 &  \textbf{0.361} & \textbf{0.393} & 0.372 & 0.400 & 0.378 & 0.407 & 0.387 & 0.412 & \underline{0.368} & 0.396 & 0.372 & \underline{0.394} & 0.374 & 0.394 & 0.412 & 0.423 \\
    & 720 &  \textbf{0.415} & \underline{0.425} & 0.427 & 0.432 & 0.433 & 0.431 & 0.464 & 0.452 & \underline{0.421} & 0.434 & 0.431 & 0.431 & \underline{0.421} & \textbf{0.423} & 0.483 & 0.463 \\
    \cmidrule(lr){2-18}
    & Avg &  \textbf{0.347} & \textbf{0.384} & 0.358 & 0.391 & 0.361 & 0.394 & 0.385 & 0.407 & \underline{0.355} & 0.389 & 0.361 & \underline{0.386} & 0.356 & \textbf{0.384} & 0.406 & 0.413 \\
    \midrule
    \multirow{5}{*}{\rotatebox{90}{ETTm2~\citep{zhou2021informer}}}
    &  96 &  \underline{0.174} & \textbf{0.259} & 0.182 & 0.268 & 0.178 & 0.261 & 0.187 & 0.270 & 0.178 & 0.268 & 0.178 & 0.265 & \textbf{0.172} & \underline{0.260} & 0.189 & 0.273 \\
    & 192 &  \textbf{0.234} & \textbf{0.299} & 0.245 & 0.310 & 0.243 & 0.304 & 0.254 & 0.317 & \underline{0.235} & 0.306 & 0.237 & 0.306 & 0.239 & \underline{0.302} & 0.253 & 0.315 \\
    & 336 &  \textbf{0.286} & \textbf{0.335} & 0.300 & 0.347 & 0.293 & 0.342 & 0.323 & 0.358 & 0.291 & 0.348 & 0.291 & 0.351 & \underline{0.289} & \underline{0.337} & 0.328 & 0.359 \\
    & 720 &  \textbf{0.372} & \textbf{0.392} & 0.377 & 0.398 & 0.379 & 0.402 & 0.432 & 0.421 & 0.385 & 0.406 & 0.403 & 0.435 & \underline{0.376} & \underline{0.397} & 0.415 & 0.418 \\
    \cmidrule(lr){2-18}
    & Avg &  \textbf{0.267} & \textbf{0.321} & 0.276 & 0.331 & 0.273 & 0.327 & 0.299 & 0.342 & 0.272 & 0.332 & 0.277 & 0.339 & \underline{0.269} & \underline{0.324} & 0.296 & 0.341 \\
    \midrule
    \multirow{5}{*}{\rotatebox{90}{Weather~\citep{wu2021autoformer}}} 
    &  96 & 0.156 & 0.206 & 0.154 & 0.205 & \textbf{0.149} & \underline{0.202} & 0.183 & 0.234 & 0.159 & 0.209 & 0.172 & 0.234 & \underline{0.150} & \textbf{0.201} & 0.173 & 0.234 \\
    &  192 & 0.205 & 0.253 & 0.204 & 0.254 & \textbf{0.195} & \underline{0.250} & 0.432 & 0.435 & 0.203 & 0.259 & 0.216 & 0.269 & \underline{0.196} & \textbf{0.248} & 0.229 & 0.275 \\
    &  336 & 0.260 & 0.295 & 0.260 & 0.297 & \underline{0.253} & \underline{0.291} & 0.534 & 0.563 & 0.256 & 0.293 & 0.263 & 0.309 & \textbf{0.246} & \textbf{0.290} & 0.298 & 0.321 \\
    &  720 & 0.334 & 0.347 & 0.336 & 0.348 & \underline{0.325} & \underline{0.346} & 0.601 & 0.578 & 0.328 & 0.348 & 0.335 & 0.358 & \textbf{0.319} & \textbf{0.342} & 0.387 & 0.375 \\
    \cmidrule(lr){2-18}
    & Avg & 0.239 & 0.275 & 0.239 & 0.276 & \underline{0.231} & \underline{0.272} & 0.438 & 0.453 & 0.237 & 0.277 & 0.247 & 0.293 & \textbf{0.228} & \textbf{0.270} & 0.272 & 0.301 \\
    \midrule
    \multirow{5}{*}{\rotatebox{90}{ECL~\citep{wu2021autoformer}}} 
    &  96  & \textbf{0.132} & \textbf{0.229} & \underline{0.135} & \underline{0.234} & 0.138 & 0.243 & 0.173 & 0.258 & 0.139 & 0.243 & 0.140 & 0.243 & \textbf{0.132} & 0.240 & 0.173 & 0.278 \\
    &  192  & \textbf{0.151} & \textbf{0.247} & 0.155 & \underline{0.253} & 0.163 & 0.265 & 0.284 & 0.367 & 0.160 & 0.263 & \underline{0.154} & 0.256 & \underline{0.154} & 0.254 & 0.183 & 0.285 \\
    &  336  & \textbf{0.168} & \textbf{0.265} & 0.174 & \underline{0.271} & 0.186 & 0.295 & 0.367 & 0.431 & 0.185 & 0.297 & 0.175 & 0.275 & \underline{0.173} & 0.274 & 0.194 & 0.305 \\
    &  720  & \textbf{0.203} & \textbf{0.295} & \underline{0.207} & \underline{0.302} & 0.258 & 0.354 & 0.442 & 0.487 & 0.264 & 0.364 & 0.211 & 0.312 & 0.226 & 0.325 & 0.223 & 0.325 \\
    \cmidrule(lr){2-18}
    & Avg  & \textbf{0.164} & \textbf{0.259} & \underline{0.168} & \underline{0.265} & 0.186 & 0.289 & 0.317 & 0.386 & 0.187 & 0.292 & 0.170 & 0.272 & 0.171 & 0.273 & 0.193 & 0.298 \\
    \midrule
    \multirow{5}{*}{\rotatebox{90}{Solar~\citep{lai2018modeling}}}  
    &  96  & \underline{0.173} & \textbf{0.224} & \textbf{0.171} & \underline{0.225} & 0.213 & 0.276 & 0.234 & 0.281 & 0.194 & 0.268 & 0.189 & 0.254 & 0.182 & 0.243 & 0.190 & 0.267 \\
    &  192  & \underline{0.195} & \textbf{0.242} & \textbf{0.194} & \underline{0.243} & 0.237 & 0.302 & 0.382 & 0.443 & 0.221 & 0.298 & 0.213 & 0.269 & 0.203 & 0.258 & 0.201 & 0.270 \\
    &  336 & \underline{0.216} & \underline{0.257} & \textbf{0.215} & \textbf{0.250} & 0.254 & 0.321 & 0.453 & 0.543 & 0.254 & 0.323 & 0.231 & 0.287 & 0.224 & 0.276 & 0.221 & 0.301 \\
    &  720  & \underline{0.245} & \underline{0.275} & \textbf{0.231} & \textbf{0.268} & 0.289 & 0.373 & 0.534 & 0.618 & 0.298 & 0.367 & 0.248 & 0.302 & 0.246 & 0.315 & 0.254 & 0.324 \\
    \cmidrule(lr){2-18}
    & Avg & \underline{0.207} & \underline{0.250} & \textbf{0.203} & \textbf{0.247} & 0.248 & 0.318 & 0.401 & 0.471 & 0.242 & 0.314 & 0.220 & 0.278 & 0.214 & 0.273 & 0.217 & 0.291 \\
    \bottomrule
  \end{tabular}}
  \end{small}
\end{table*}

\subsection{Dynamic Mixture-of-Experts Framework Analysis}  
\label{sec:moe_analysis}  
Table~\ref{tab:forecasting_promotion} reveals that the dynamic mixture-of-experts framework demonstrates dual advantages in temporal forecasting. First, it exhibits universal adaptability across diverse model architectures (LLaMA~\citep{dubey2024llama}, OPT~\citep{zhang2022opt}, GPT2~\citep{radford2019language}) and data characteristics. With the dynamic mixture-of-experts framework, the prediction accuracy is highly versatile across models of different sizes or data sets of different types. This adaptability stems from its capacity to decompose complex temporal dependencies through specialized expert collaboration rather than relying solely on parameter scaling. Second, the framework establishes a lightweight paradigm for resource-efficient deployment. Smaller models integrated with the dynamic mixture-of-experts framework achieve competitive performance comparable to larger baselines by selectively activating domain experts, effectively balancing capacity and computational cost. This synergy of generalization and efficiency positions the dynamic mixture-of-experts framework as a strategic enabler for scalable time-series intelligence.

\vspace{-9pt}
\begin{table}[htbp]
  \caption{Performance promotion obtained by our Dynamic Mixture-of-Experts Framework. We report the average performance and the relative MSE reduction (Promotion), where \textbf{bold numbers} indicate the best performance of different LLMs as backbone in different data sets.The \textbf{Original} entries denote the baseline SMETimes performance without employing the dynamic mixture-of-experts framework, providing direct ablation comparisons.} 
  \label{tab:forecasting_promotion}
  \vskip 0.05in
  \centering
  \footnotesize
  \setlength{\tabcolsep}{1.5pt} 
  \renewcommand{\arraystretch}{1.0}
  \resizebox{\textwidth}{!}{
  \begin{tabular}{@{}c|c|cc|cc|cc|cc|cc@{}}
    \toprule
    \multicolumn{2}{c|}{\multirow{2}{*}{\raisebox{1.0ex}{Experiment}}} & 
    \multicolumn{2}{c|}{LLaMA-3B~\citep{dubey2024llama}} &
    \multicolumn{2}{c|}{LLaMA-1B~\citep{dubey2024llama}} &
    \multicolumn{2}{c|}{OPT-2.7B~\citep{zhang2022opt}} &
    \multicolumn{2}{c|}{OPT-1.3B~\citep{zhang2022opt}} &
    \multicolumn{2}{c}{GPT2-124M~\citep{radford2019language}} \\
    \cmidrule(lr){3-4} \cmidrule(lr){5-6}\cmidrule(lr){7-8} \cmidrule(lr){9-10} \cmidrule(lr){11-12}  
    \multicolumn{2}{c|}{\raisebox{1.5ex}{Setting}}  & MSE & MAE  & MSE & MAE  & MSE & MAE  & MSE & MAE & MSE & MAE\\
    \toprule
    \multirow{3}{*}{\raisebox{-1ex}{Weather~\citep{wu2021autoformer}}} & Original & 0.269 & 0.321 & 0.276 & 0.329 & 0.258 & 0.312 & 0.263 & 0.324 & 0.279 & 0.341\\
    &\textbf{+MoE} & \textbf{0.239} & \textbf{0.275} & \textbf{0.235} & \textbf{0.273} & \textbf{0.235} & \textbf{0.272} & \textbf{0.239} & \textbf{0.275} & \textbf{0.243} & \textbf{0.281}\\
    \cmidrule(lr){2-12}
    &Promotion & +11.1\% & +14.3\% & +14.9\% & +17.0\% & +8.9\% & +12.8\% & +9.1\% & +15.1\% & +12.9\% & +17.6\%\\
    \midrule
    \multirow{3}{*}{\raisebox{-1ex}{ECL~\citep{wu2021autoformer}}} & Original & 0.178 & 0.276& 0.186 & 0.281 & 0.175 & 0.271  & 0.185 & 0.279 & 0.193 & 0.293\\
    &\textbf{+MoE} & \textbf{0.159} & \textbf{0.254} & \textbf{0.164} & \textbf{0.259} & \textbf{0.158} & \textbf{0.254} & \textbf{0.159} & \textbf{0.255} & \textbf{0.174} & \textbf{0.266}\\
    \cmidrule(lr){2-12}
    &Promotion & +10.7\% & +8.0\% & +11.8\% & +7.8\%  & +9.7\% & +6.3\% & +14.1\% & +8.6\% & +9.8\% & +9.2\%\\
    \midrule
    \multirow{3}{*}{\raisebox{-1ex}{Solar~\citep{lai2018modeling}}} & Original & 0.234 & 0.281 & 0.245 & 0.287 & 0.232 & 0.297 & 0.245 & 0.287 & 0.255 & 0.312\\
    &\textbf{+MoE} & \textbf{0.207} & \textbf{0.250} & \textbf{0.205} & \textbf{0.252} & \textbf{0.208} & \textbf{0.262} & \textbf{0.208} & \textbf{0.259} & \textbf{0.219} & \textbf{0.278}\\
    \cmidrule(lr){2-12}
    &Promotion & +11.5\% & +11.0\% & +16.3\% & +12.2\% & +10.3\% & +11.8\% & +15.1\% & +9.8\% & +14.1\% & +10.9\%\\
    \bottomrule
  \end{tabular}}
\end{table}

\subsection{Ablation Studies}  
Systematic ablation studies validate the necessity of the core components of SMETimes, as shown in Table~\ref{tab:ablation}. Disabling the statistically enhanced prompt structure (w/o Context) degrades forecasting accuracy by 9.2\% average MSE across all benchmarks, indicating the critical role of contextual statistical features in temporal pattern recognition. The adaptive fusion embedding structure is essential for cross-modal alignment—its removal (w/o Fusion) causes 15.7\% overall performance deterioration, with particularly severe degradation on the Weather~\citep{wu2021autoformer} dataset (18.2\% MSE increase). Most crucially, deactivation of the MoE structure (w/o MoE) results in 12.4\% average accuracy loss, peaking at 15.1\% MSE reduction on Solar~\citep{lai2018modeling} predictions, which demonstrates the effectiveness of expert specialization in handling heterogeneous temporal patterns. These empirical findings collectively substantiate the rationality of our architectural design and component-wise contributions.

\begin{table*}[htbp]
  \caption{Ablation of method designs. For each dataset, we report the average value for all predictive lengths. \textbf{Bold numbers} represent the best performance for a particular combination of modules in the same data set. Underlined numbers represent the second-best performance.}
  \label{tab:ablation}
  \centering
  \footnotesize
  \setlength{\tabcolsep}{2.5pt}
  \renewcommand{\arraystretch}{0.9}
  \begin{tabular}{c|ccc|cc|cc|cc|cc}
    \toprule
    \raisebox{-2ex}[0pt][0pt]{
    \begin{tabular}{c}
        Experiment \\ 
        Setting
    \end{tabular}} & \multicolumn{3}{c}{\raisebox{0ex}[0pt][0pt]{Module}} 
    & \multicolumn{2}{|c|}{ETTh1~\citep{zhou2021informer}} 
    & \multicolumn{2}{c|}{ETTh2~\citep{zhou2021informer}} 
    & \multicolumn{2}{c|}{ETTm1~\citep{zhou2021informer}}
    & \multicolumn{2}{c}{ETTm2~\citep{zhou2021informer}}\\
    \cmidrule(lr){2-4} \cmidrule(lr){5-6} \cmidrule(lr){7-8} 
    \cmidrule(lr){9-10} \cmidrule(lr){11-12}
    & Context & Fusion & MoE & MSE & MAE & MSE & MAE & MSE & MAE & MSE & MAE \\
    \midrule
    Original & \ding{51} & \ding{51} & \ding{51} 
    & \textbf{0.387} & \textbf{0.420} & \textbf{0.349} & \textbf{0.399} & \textbf{0.347} & \textbf{0.384} & \textbf{0.267} & \textbf{0.321} \\
    w/o Context & \ding{51} & \ding{55} & \ding{55} 
    & \underline{0.391} & \underline{0.427} & \underline{0.356} & \underline{0.402} & \underline{0.349} & 0.387 & \underline{0.269} & \underline{0.327} \\
    w/o Fusion & \ding{55} & \ding{51} & \ding{55} 
    & 0.428 & 0.452 & 0.398 & 0.437 & 0.359 & \underline{0.386} & 0.287 & 0.347 \\
    w/o MoE & \ding{55} & \ding{55} & \ding{51} 
    & 0.403 & 0.442 & 0.382 & 0.423 & 0.356 & 0.399 & 0.289 & 0.342 \\
    \bottomrule
  \end{tabular}
\end{table*}

\subsection{Hyperparameter Sensitivity} 
\label{sec:sensitivity}
\begin{figure}[htbp]
  \centering
  \begin{subfigure}[b]{0.32\textwidth}
    \includegraphics[width=\textwidth]{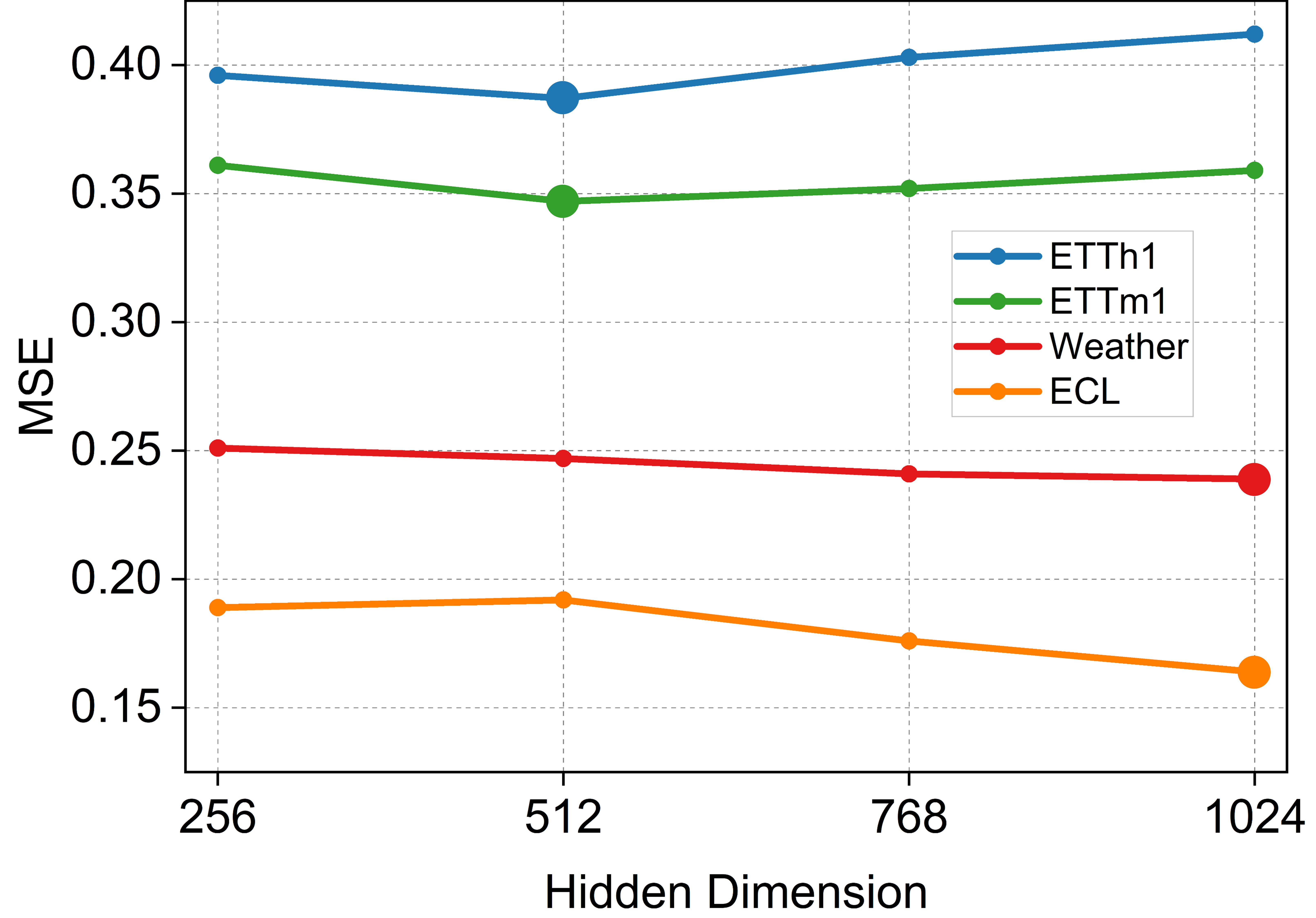}
    \subcaption{(a): Hidden Dimension}
    \label{fig:parametersub1}
  \end{subfigure}
  \hfill
  \begin{subfigure}[b]{0.32\textwidth}
    \includegraphics[width=\textwidth]{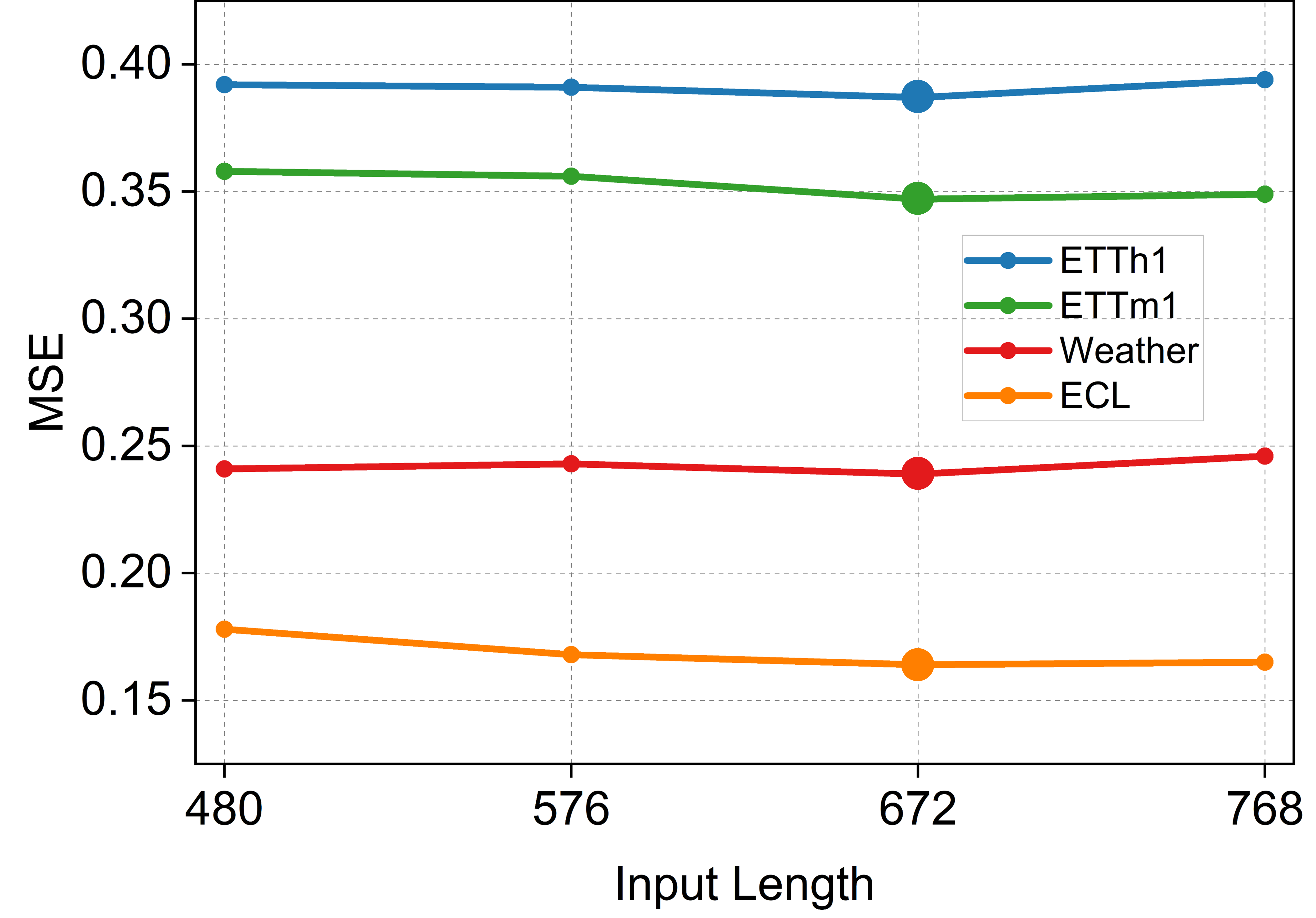}
    \subcaption{(b): Input Length}
    \label{fig:parametersub2}
  \end{subfigure}
  \hfill
  \begin{subfigure}[b]{0.32\textwidth}
    \includegraphics[width=\textwidth]{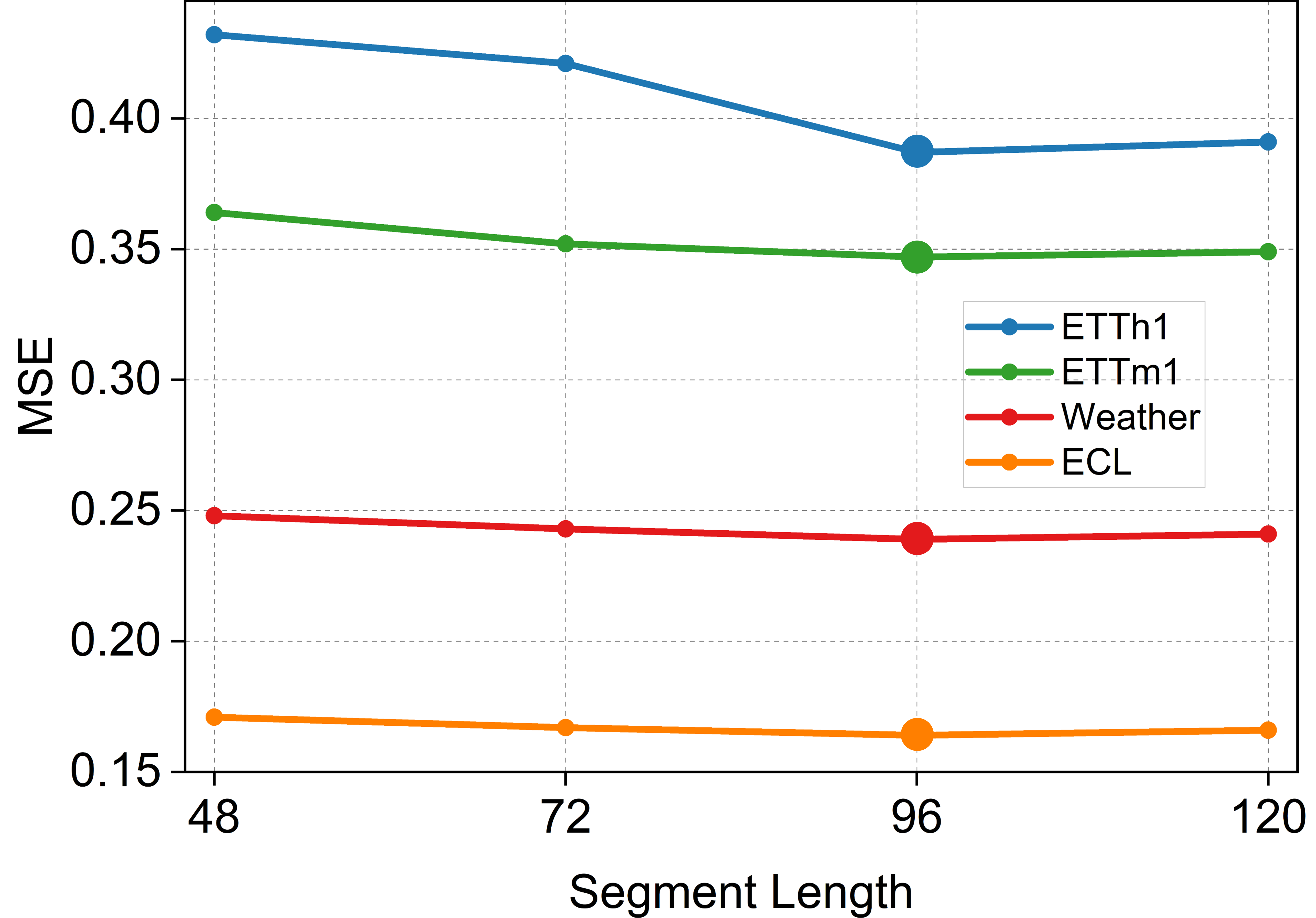}
    \subcaption{(c): Segment Length}
    \label{fig:parametersub3}
  \end{subfigure}
  
  \caption{Hyperparameter sensitivity of SMETimes. Each curve presents a specific dataset.}
  \label{fig:hyperparameter}
\end{figure}

As shown in Fig.~\ref{fig:hyperparameter}, comprehensive analyzes across representative datasets (ETTh1~\citep{zhou2021informer}, ETTm1~\citep{zhou2021informer}, Weather~\citep{wu2021autoformer}, ECL~\citep{wu2021autoformer}) reveal the stable performance of the proposed SMETimes under varying configurations. The \textbf{bolded} points represent the best results in this dataset. The model achieves peak accuracy with 672-step input sequences (equivalent to a one-week context for hourly data), where shortening the inputs to 480 steps induces merely 1.3\% MSE degradation. Temporal segmentation analysis identifies 96-step windows as optimal to align language model processing with periodic patterns, while hidden dimension studies demonstrate that 512-channel projections optimally balance computational efficiency and representational capacity: Expanding to 1024 dimensions yields diminishing returns ($<$0.9\% accuracy gain despite 2.1$\times$ computation overhead). In particular, the framework exhibits robust generalization, maintaining performance within $\pm$5\% of optimal MSE across all hyperparameter combinations tested, confirming its reliability for practical deployment.

\subsection{Showcases}
\begin{figure}[htbp]
  \centering
  \begin{subfigure}[b]{0.32\textwidth}
    \includegraphics[width=\textwidth]{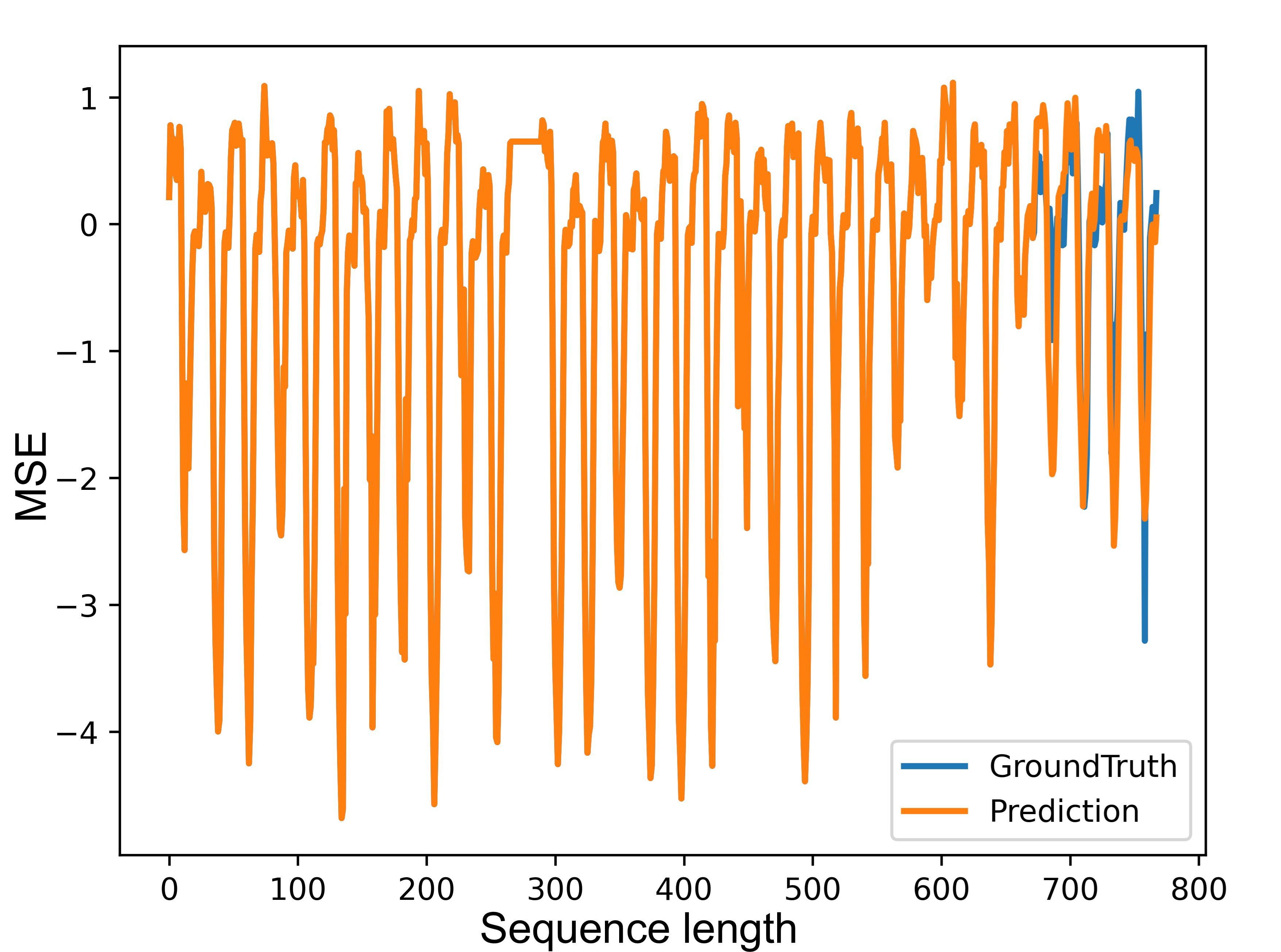}
    \subcaption{(a1): SMETimes(ours)}
    \label{fig:case1sub1}
  \end{subfigure}
  \hfill
  \begin{subfigure}[b]{0.32\textwidth}
    \includegraphics[width=\textwidth]{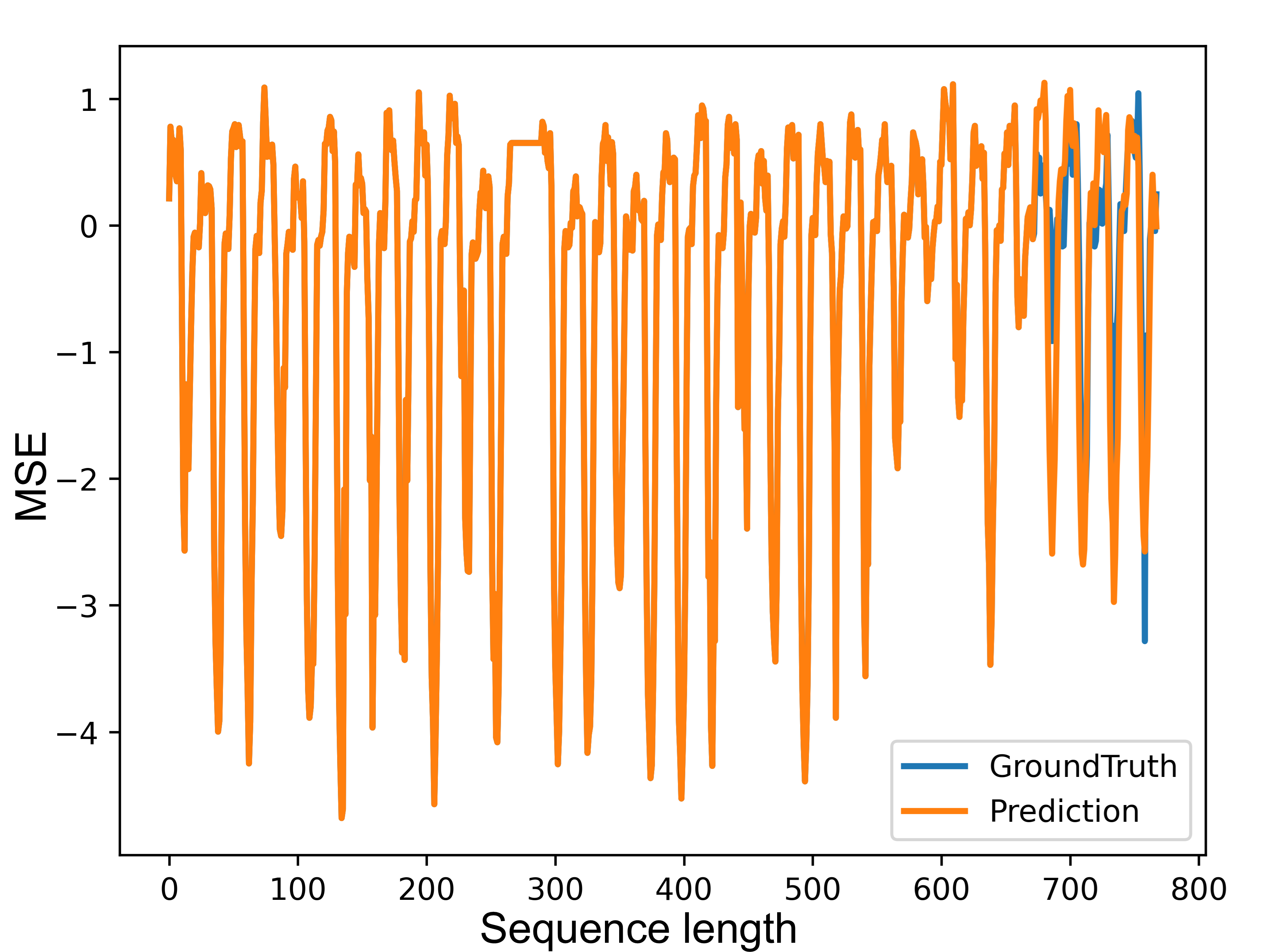}
    \subcaption{(b1): AutoTimes~\citep{liu2025autotimes}}
    \label{fig:case1sub2}
  \end{subfigure}
  \hfill
  \begin{subfigure}[b]{0.32\textwidth}
    \includegraphics[width=\textwidth]{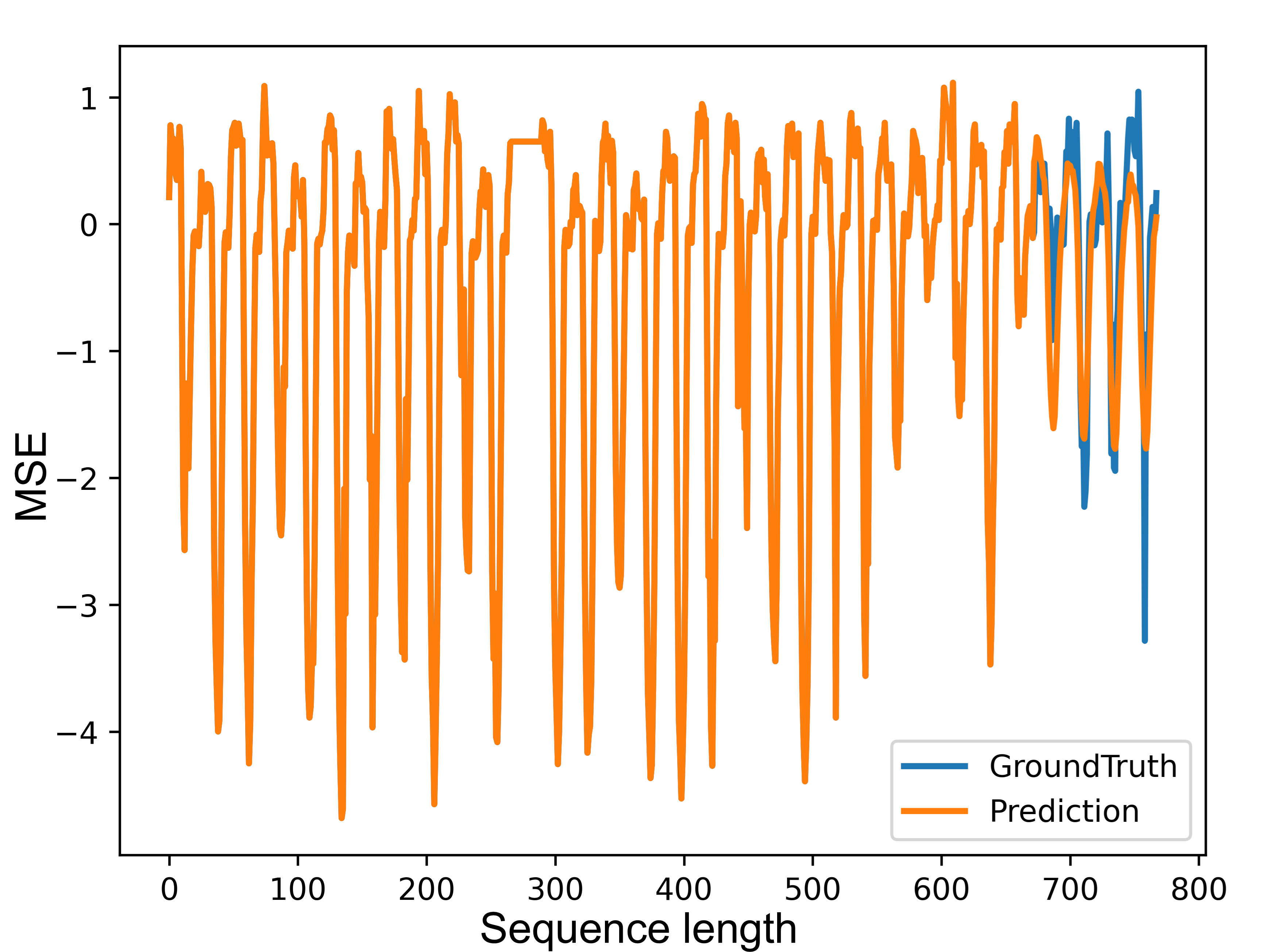}
    \subcaption{(c1): Time-LLM~\citep{jin2023time}}
    \label{fig:case1sub3}
  \end{subfigure}

  \vspace{10pt}
  \begin{subfigure}[b]{0.32\textwidth}
    \includegraphics[width=\textwidth]{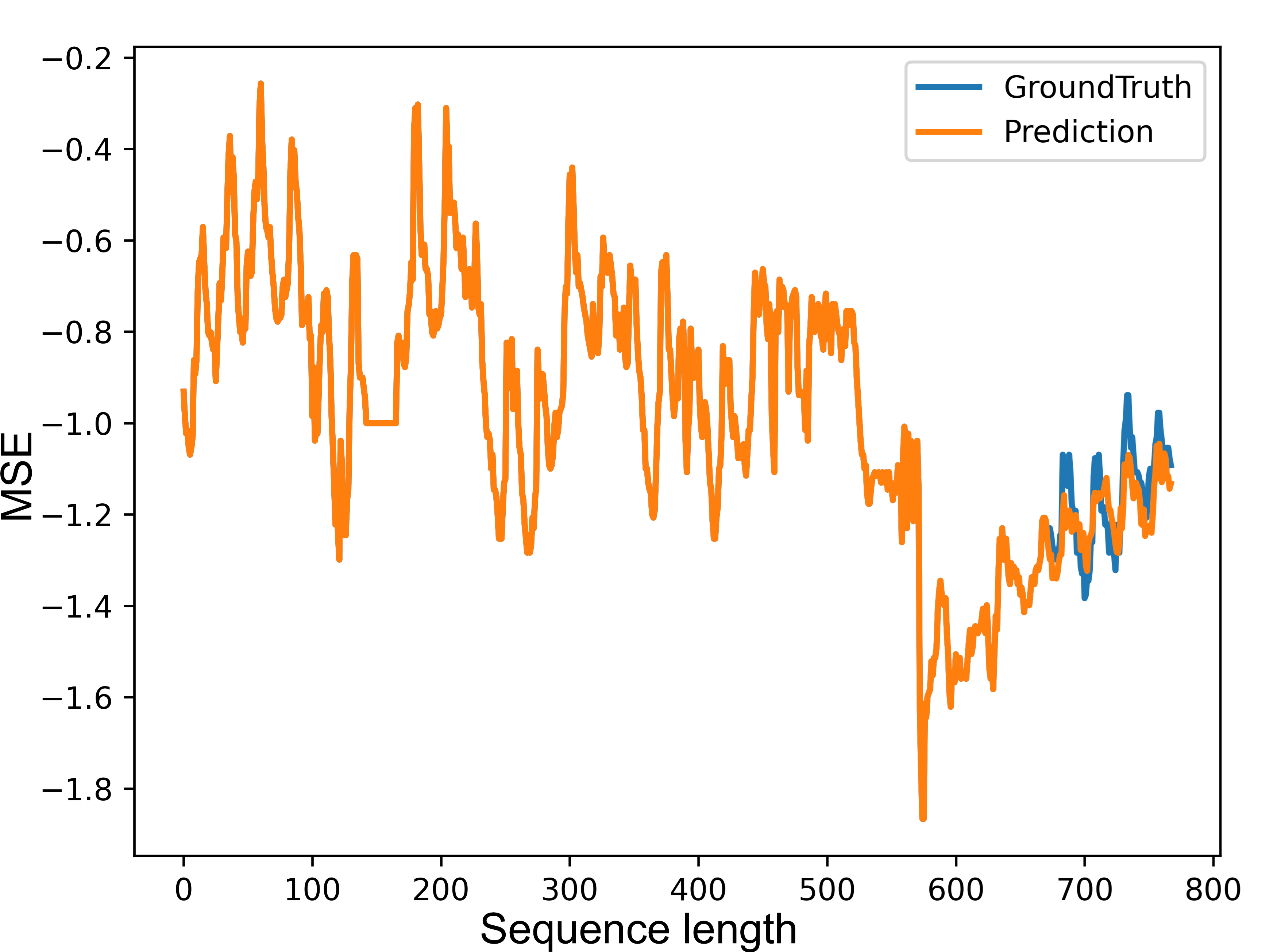}
    \subcaption{(a2): SMETimes(ours)}
    \label{fig:case1sub4}
  \end{subfigure}
  \hfill
  \begin{subfigure}[b]{0.32\textwidth}
    \includegraphics[width=\textwidth]{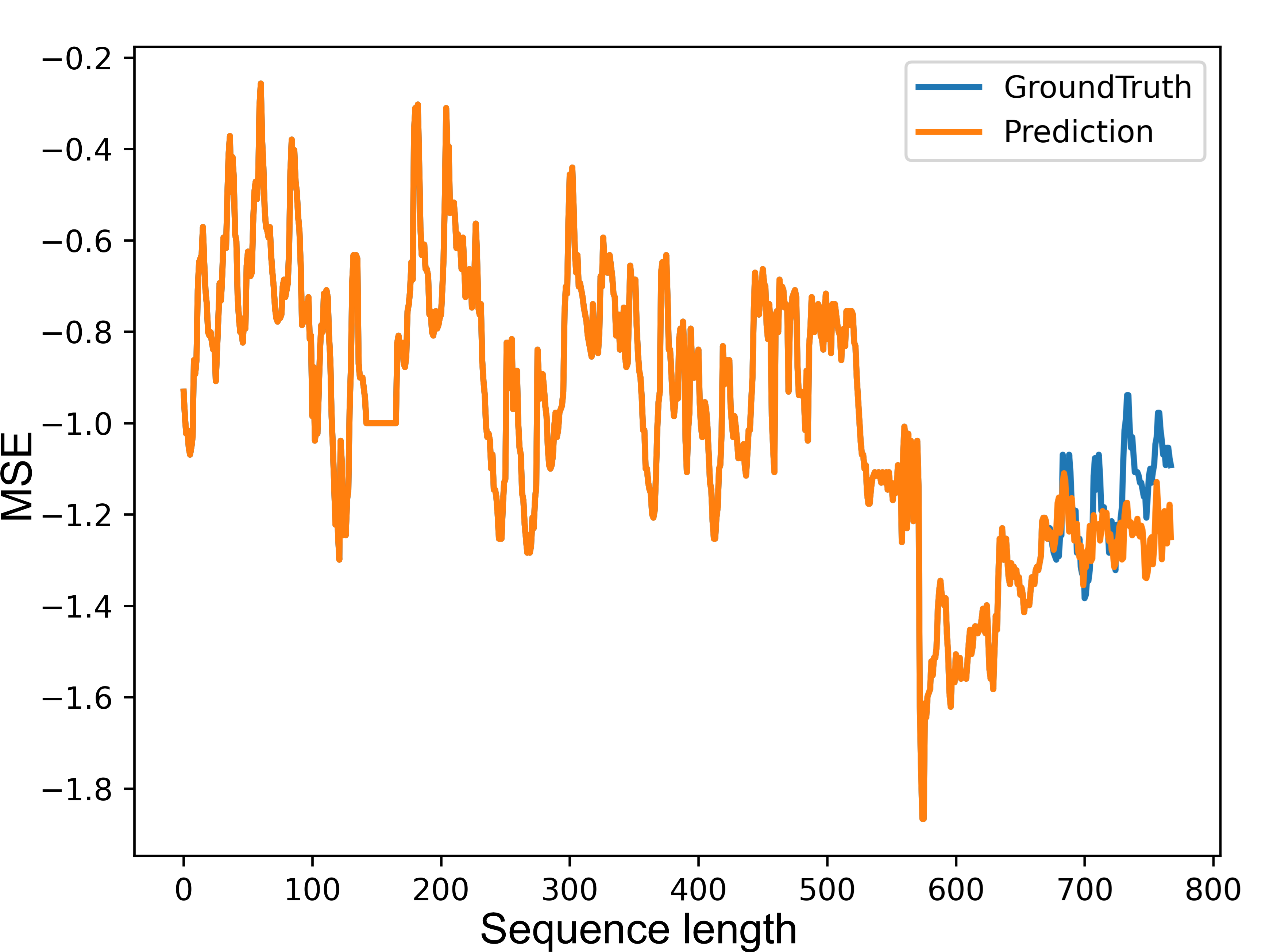}
    \subcaption{(b2): AutoTimes~\citep{liu2025autotimes}}
    \label{fig:case1sub5}
  \end{subfigure}
  \hfill
  \begin{subfigure}[b]{0.32\textwidth}
    \includegraphics[width=\textwidth]{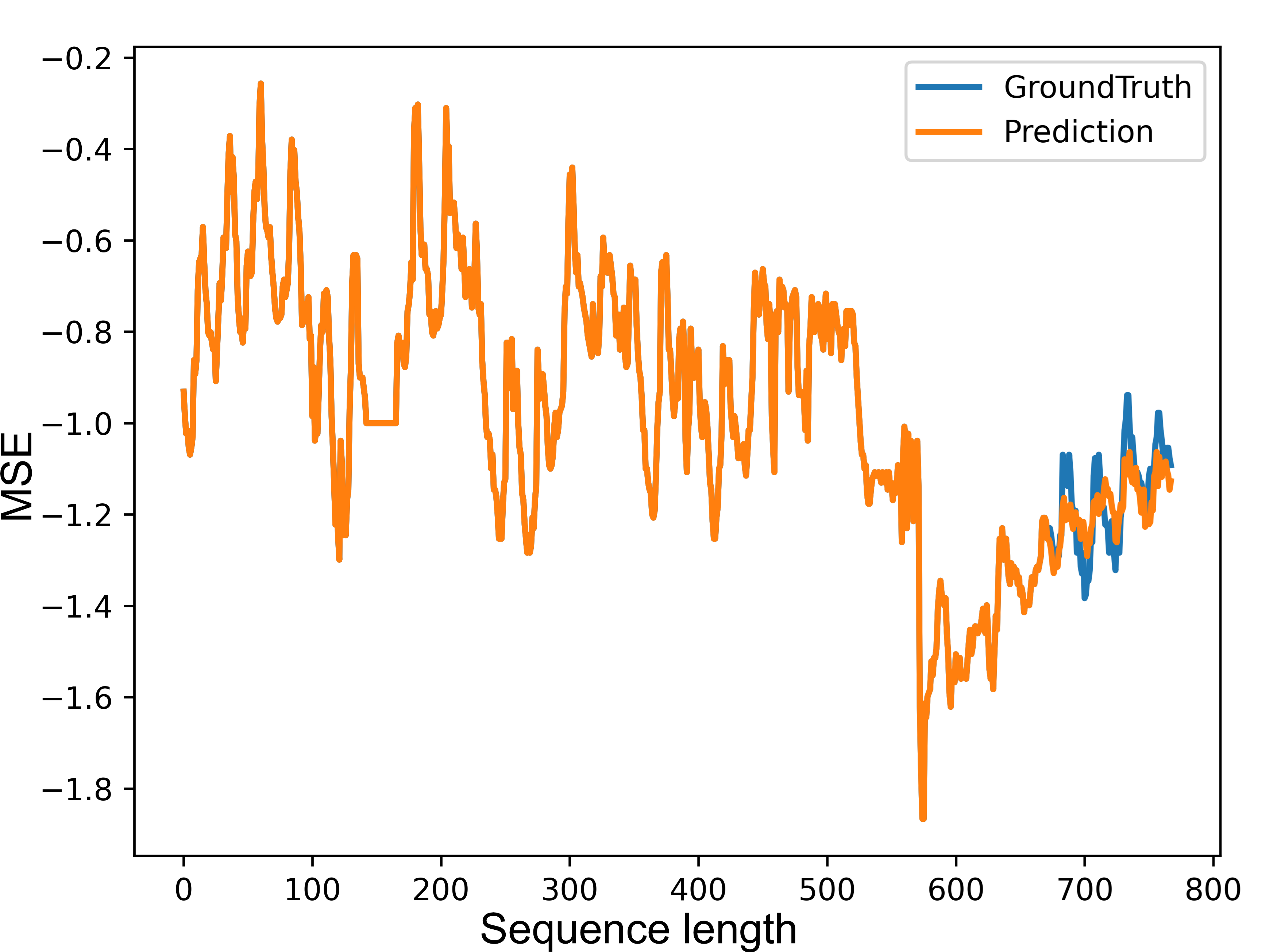}
    \subcaption{(c2): Time-LLM~\citep{jin2023time}}
    \label{fig:case1sub6}
  \end{subfigure}

  \caption{Long-term forecasting cases from ETTh1~\citep{zhou2021informer} by different models under the input-672-predict-96 settings. \textcolor{blue}{Blue} lines are the ground truths and \textcolor{orange}{orange} lines are the model predictions.}
  \label{fig:case}
\end{figure}

As demonstrated in Fig.~\ref{fig:case}, we present a comparative analysis of long-term forecasting performance under the input-672-predict-96 configuration using the ETTh1~\citep{zhou2021informer} dataset. The subgraphs Fig.~\ref{fig:case}(a1) and Fig.~\ref{fig:case}(a2) demonstrate that our SMETimes can effectively capture temporal patterns with more stable predictions, indicating its strong capability to learn time series dynamics. In contrast, the subgraphs Fig.~\ref{fig:case}(b1), Fig.~\ref{fig:case}(b2) reveal that AutoTimes~\citep{liu2025autotimes} tends to produce more aggressive predictions with higher oscillation amplitudes, while Fig.~\ref{fig:case}(c1) and Fig.~\ref{fig:case}(c2) illustrate Time-LLM's~\citep{jin2023time} conservative prediction strategy characterized by noticeable lagging effects. In this evaluation, SMETimes exhibits improved prediction accuracy compared to contemporary LLM-based time series forecasting models, including AutoTimes~\citep{liu2025autotimes} and Time-LLM~\citep{jin2023time}.

\section{Limitation}\label{sec:limitations}
Our model suffers from the degradation of accuracy in longer-horizon forecasting due to the capacity constraints of Small Language Models (SLMs). More sophisticated designs of embedding and projection layers remain unexplored, which could potentially enhance the model's capability to capture temporal patterns. Additionally, training efficiency could be further optimized using advanced techniques such as dynamic batching or mixed-precision training, as the current computational overhead still poses challenges for resource-constrained scenarios. These promising directions constitute our immediate research agenda to improve both performance and practicality.

\section{Acknowledge}\label{sec:acknowledge}  
This work was supported by the Natural Science Foundation of Guangdong Province (No. 2023A1515010673), in part by the Shenzhen Science and Technology Innovation Bureau key project (No. JSGG20220831110400001, No. CJGJZD20230724093303007, KJZD20240903101259001), in part by Shenzhen Medical Research Fund (No. D2404001), in part by Shenzhen Engineering Laboratory for Diagnosis \& Treatment Key Technologies of Interventional Surgical Robots (XMHT20220104009), and the Key Laboratory of Biomedical Imaging Science and System, CAS, for the Research platform support.

\section{Conclusion}\label{sec:conclusion}  
In this work, we proposed the SMETimes, which establishes SLMs as efficient time series forecasters through three innovations. The statistically enhanced prompt structure bridges numerical temporal signals, the adaptive fusion embedding structure aligns continuous patterns, and the dynamic mix-of-experts structure leverages the SLM efficiency. Our 3B model outperforms the 7B LLMs by 6.9\% MSE with 3.8$\times$ faster inference, achieving SOTA on five benchmarks. Ablations validate critical components (15.7\% error reduction from adaptive fusion embedding structure), while maintaining stability across configurations. These prospective avenues constitute our primary research agenda that aims to enhance both the performance and practical applicability of the model.

\bibliography{sn-bibliography}

\end{document}